\documentclass[lettersize,journal]{IEEEtran}
\usepackage{amsmath,amsfonts}
\usepackage{algorithmic}
\usepackage{algorithm}
\usepackage{array}
\usepackage[caption=false,font=normalsize,labelfont=sf,textfont=sf]{subfig}
\usepackage{textcomp}
\usepackage{stfloats}
\usepackage{url}
\usepackage{verbatim}
\usepackage{graphicx}
\usepackage{cite}
\usepackage{multicol}
\usepackage{multirow}

\usepackage{makecell}
\usepackage{xcolor}
\usepackage{booktabs}
\usepackage{placeins}

\begin{document}

\title{GBSVR: Granular Ball Support Vector Regression}

\author{Reshma Rastogi,~\IEEEmembership{Senior Member, ~IEEE},
    Ankush Bisht, 
    Sanjay Kumar, 
    and Suresh Chandra
    \thanks{Reshma Rastogi and Ankush Bisht are with the MLSI Lab, Department of Computer Science, South Asian University, New Delhi, India (e-mail: reshma.khemchandani@sau.ac.in; ankushbisht72@gmail.com)}
    \thanks{Sanjay Kumar is with Department of Computer Science, Deshbandhu College, University of Delhi, New Delhi, India (e-mail: skumar5@db.du.ac.in)}
    \thanks{Suresh Chandra is Professor Emeritus with the Department of Mathematics, Indian Institute of Technology Delhi, New Delhi, India}

\thanks{Manuscript received March, 2025.}
}

\markboth{Journal of \LaTeX\ Class Files,~Vol.~14, No.~8, March~2025}%
{Shell \MakeLowercase{\textit{et al.}}: A Sample Article Using IEEEtran.cls for IEEE Journals}


\maketitle

\begin{abstract}

Support Vector Regression (SVR) and its variants are widely used to handle regression tasks, however, since their solution involves solving an expensive quadratic programming problem, it limits its application, especially when dealing with large datasets. Additionally, SVR  uses an $\epsilon$-insensitive loss function which is sensitive to outliers and therefore can adversely affect its performance. We propose Granular Ball Support Vector Regression (GBSVR) to tackle problem of regression by using granular ball concept. These balls are useful in simplifying complex data spaces for machine learning tasks, however, to the best of our knowledge, they have not been sufficiently explored for regression problems. Granular balls group the data points into balls based on their proximity and reduce the computational cost in SVR by replacing the large number of data points with far fewer granular balls. This work also suggests a discretization method for continuous-valued attributes to facilitate the construction of granular balls.
The effectiveness of the proposed approach is evaluated on several benchmark datasets and it outperforms existing state-of-the-art approaches. 
\end{abstract}


\section{Introduction}
\IEEEPARstart{R}{egression} is a fundamental task in machine learning and statistical modeling, where the goal is to predict a continuous target variable based on a given set of input features \cite{darlington2016regression, gunst2018regression, zaki2020data}. Ridge regression, lasso, and SVR are widely used regression techniques \cite{o2016statistical,drucker1996support}. SVR has demonstrated strong performance in various real-world applications \cite{muchtadi2024support} due to its ability to handle non-linear relationships and deliver high prediction accuracy \cite{bi2003geometric, vapnik2013nature}. There are several extensions of the SVR model that are often based on a variety of loss functions \cite{rastogi2017norm1, rastogi2017nu,zhang2024multi} or applications \cite{gao2019end,gu2023incremental}. However, despite its advantages, SVR faces significant challenges when applied to large datasets, mainly due to its high computational cost $O(m^3)$, where $m$ is the number of training samples. The computational overhead, in terms of both space and time, arises from the need to solve a quadratic programming problem along with the storage of the kernel matrix in proportion to the number of data points and the dimensionality of the input space \cite{yang2025flexible}. This limitation restricts the scalability of SVR in big data scenarios and impacts its practicality in real-time applications. 
SVR is also sensitive to outliers and noisy data points \cite{chuang2002robust} and poses a major challenge. 
These limitations necessitate alternative strategies that can reduce computational overhead and improve model robustness without sacrificing prediction accuracy. One promising approach is to represent the data in a structured and smooth manner rather than processing each individual point separately.

Granular computing is a computing paradigm that can potentially process large-scale data with imprecise information using varied-sized granular balls, which are crude yet effective representations of the underlying data \cite{zadeh2000fuzzy}. By capturing data in a structured, coarse-to-fine manner, granular-ball computing offers several advantages. It improves robustness by mitigating the impact of noise, and improves interpretability by highlighting relationships in groups of related data rather than treating them as isolated points. These characteristics make it beneficial for handling large-scale and high-dimensional datasets \cite{xia2019granular}.

To address issues associated with SVR, this paper introduces Granular Ball Support Vector Regression (GBSVR), an approach that integrates the concept of granular balls into the SVR framework. The concept of Granular Balls is being utilized in various machine learning algorithms when dealing with large datasets \cite{xie2023efficient,xia2024gbsvm,xia2024granular}. Granular balls reduce computational complexity, and their structure helps mitigate the impact of outliers by focusing on regions of high data density and reducing noise interference \cite{bai2023granular,xia2021granular}. Through the integration of granular balls, GBSVR improves the efficiency of the regression model while maintaining and often improving prediction accuracy. This approach makes it relevant for real-world scenarios where datasets are often massive, noisy, and computationally expensive to process using traditional methods.

The key contributions of this paper are as follows:
\begin{itemize}
    \item We propose GBSVR, a novel approach that leverages granular computing to lower computational cost and integrates with statistical learning to improve the scalability and performance of SVR.
    
    \item We introduce a new discretization method for the continuous-valued prediction variable to facilitate construction of the granular regression balls. 
    \item We evaluate the proposed approach on several benchmark datasets, including time series data, demonstrating that GBSVR outperforms existing state-of-the-art methods in terms of both accuracy and computational cost.
\end{itemize}

The paper is organized as follows: Section 2 covers related work, including SVR and granular ball construction. Section 3 details the proposed GBSVR approach. Section 4 presents experiments and comparisons. Section 5 concludes with a summary and future directions.

\section{Related Work}
In this section, we discuss Support Vector Regression (SVR) model, its limitations, the granular ball concept, and its role in improving traditional learning models.
\subsection{Support Vector Regression (SVR)}
A common implementation of Support Vector Regression (SVR) is the $\epsilon$-insensitive SVR that introduces a tolerance margin ($\epsilon$) to approximate target values and ignore errors smaller than $\epsilon$. By solving a quadratic programming problem, SVR identifies a regressor ($f(x)$) that minimizes the $\epsilon$-insensitive Hinge loss on a given set of data points $D=(X,Y)=\{(x_i,y_i), i=1,2,\dots,m\}$, where $x_i \in \mathcal{R}^l$, and $y_i \in \mathcal{R}$. 


The goal of SVR is to find a regressor $f(x)=w \cdot x+b$~(linear case) that trades-off between model complexity and prediction performance. The optimization problem for the linear case is formulated as:

\begin{eqnarray}
\min_{w, b, \xi, \xi^*} &\frac{1}{2} \|w\|^2 + C \sum_{i=1}^m (\xi_i + \xi_i^*),
\nonumber\\
\mbox{subject to} &\nonumber\\
&\begin{aligned}
    & y_i - w \cdot x_i - b \leq \epsilon + \xi_i,i=1,2,\ldots, m \\
    & w \cdot x_i + b - y_i \leq \epsilon + \xi_i^*,i=1,2,\ldots, m \\
    & \xi_i, \xi_i^* \geq 0 \quad \forall i.
\end{aligned}
\end{eqnarray}
Here, \(\|w\|^2\) is regularization term used to control the model complexity and flatness of the regression curve, $\epsilon$ is the margin of tolerance, \(C\) is trade-off parameter between model complexity and the penalty for large deviations. The slack variables \(\xi_i, \xi_i^*\) are to account for error corresponding to the data points outside the \(\epsilon\)-tube, above and below, respectively.

SVR's optimization problem has a computational complexity of $O(m^3)$ resulting in high computational time and memory requirements. Chunking and sequential minimal optimization (SMO) \cite{yong2004improved,zhou2013study} improve efficiency but struggle with scalability. Additionally, SVR is sensitive to outliers, which can skew predictions. Robust SVR variants \cite{chuang2002robust,sabzekar2021robust,ye2020robust,fu2023robust} mitigate this but add computational overhead.

\subsection{Granular Ball Computing}
Granular computing provides efficient and scalable methods using approximate solutions \cite{chen1982topological, xia2019granular, guo2021trend}. The granular ball concept has shown success in clustering \cite{xie2023efficient, zhang2024multi}, label denoising \cite{kong2022research}, and classification \cite{xia2022efficient}, but its application in regression remains unexplored \cite{xia2023granular}.

 A granular ball $B(c, r)$ represents a data subset using its center $c$ and radius $r$, where $c$ is the mean and $r$ quantifies the data spread. It is defined as:
 \begin{equation}
B(c, r) = \{x \in \mathbb{R}^l \mid d(x, c) \leq r\},
\end{equation}
where $d(x,c)$ is the Euclidean distance given by:

\begin{equation}
d(x, c) = \sqrt{\sum_{j=1}^l (x_j - c_j)^2}.
\end{equation}

The center \(c\) and radius \( r \) of a ball with \( u \) data points are computed as:
\begin{equation}
c = \frac{1}{u} \sum_{i=1}^u x_i, \quad r = \max_{x_i \in B} d(x_i, c).
\label{radius}
\end{equation}

Alternatively, \(r\) can be the mean distance, offering better robustness to outliers.

SVR research has focused on sampling \cite{hui2016heuristic} and approximation methods \cite{le2017approximation} to improve efficiency. Methods like reduced SVR \cite{shieh2010reduced} and approximate kernels \cite{li2016fast} reduce input points but trade efficiency for accuracy. Granular ball integration into SVR presents a novel approach to improve both scalability and robustness of SVR, as detailed in the proposed GBSVR.



\subsection{Key observations of Con-MGSVR}
The controllable multigranularity support vector algorithm (con-MGSVR) \cite{shao2025mgsvm} integrates granular balls with SVR but has unaddressed issues. It lacks clarity on mapping regressor values, $y$ to granular ball space and ignores the values when constructing granular balls, causing inconsistencies. The algorithm description conflicts with its equations, omitting
$||w||r$ in constraints (where $w$ and $r$ refer to model parameter and radius respectively) and inconsistently treating
$\epsilon$ as both an optimized variable and a fixed parameter. Furthermore, the claim that con-MGSVR reduces to SVR as $r$ tends to $0$ lacks a clear justification. This work addresses these issues and provides a more consistent GBSVR formulation.

\section{Proposed Model}
 In this section, we develop a novel approach that generates granular regression balls for continuous-valued attributes and integrates them with SVR to create a novel framework for regression task resulting in a Granular Ball Support Vector Regression (GBSVR), a robust approach that combines granular data representation with SVR’s prediction capabilities.
 
 
\subsection{Granular Regression Ball Generation}
    
    
    

\begin{figure}[ht]
    \centering
    \begin{minipage}{0.48\linewidth}
        \centering
        \includegraphics[width=\linewidth]{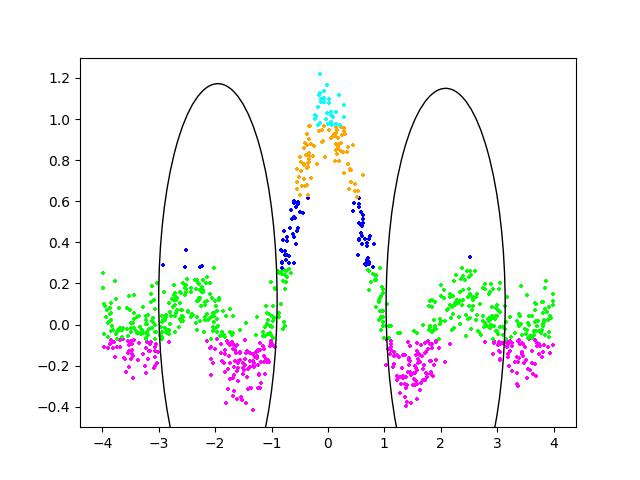}
        \text{Step 1}
    \end{minipage}
    \hfill
    \begin{minipage}{0.48\linewidth}
        \centering
        \includegraphics[width=\linewidth]{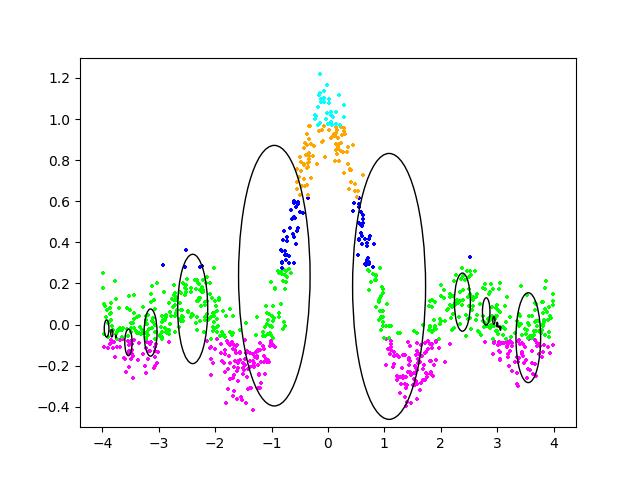}
        \text{Step 2}
    \end{minipage}
    
    \vskip 0.3cm
    
    \begin{minipage}{0.48\linewidth}
        \centering
        \includegraphics[width=\linewidth]{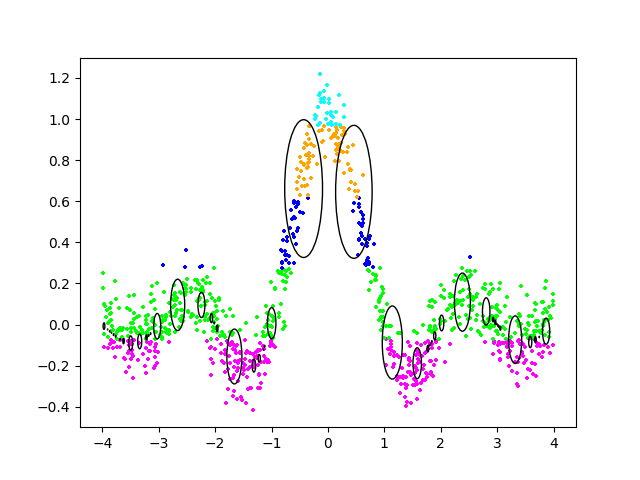}
        \text{Step 3}
    \end{minipage}
    \hfill
    \begin{minipage}{0.48\linewidth}
        \centering
        \includegraphics[width=\linewidth]{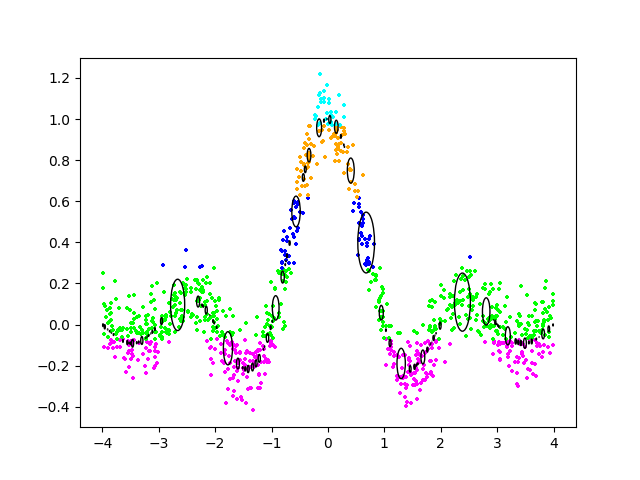}
        \text{Step 4}
    \end{minipage}
    
    \caption{Granular Regression Ball Generation Process for sinc dataset}
    \label{Steps_grid}
\end{figure}

The essence of granular regression balls lies in their ability to summarize data points into compact and meaningful clusters. For continuous variables, the radius \(r\) of a granular regression ball, as defined in (\ref{radius}), encapsulates the coverage or density of the ball. Although \(r\) can be computed in various ways, we argue that using the mean distance of all points from the center \(c\) produces granular regression balls that are more representative of the underlying data distribution. This choice reduces susceptibility to outliers and noise, compared to radii derived from maximum or minimum distances.

\begin{figure}[h]
\centering
\includegraphics[width=\textwidth]{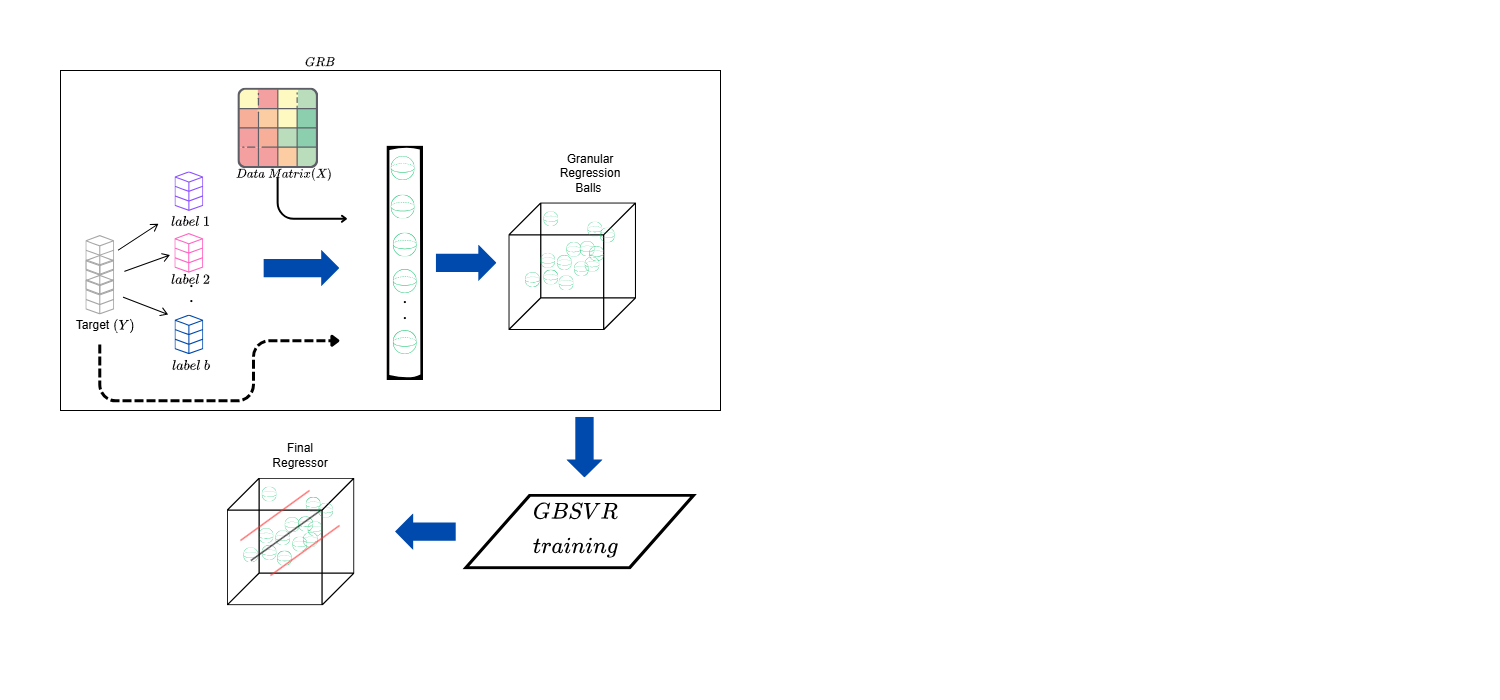}
\caption{Process of Granular Regression Ball Generation}
\label{GranularBallgen}
\end{figure}

Figs. \ref{Steps_grid} and \ref{GranularBallgen} illustrate the iterative process of generating granular regression balls. However, generating granular regression balls for regression tasks introduces unique challenges. In classification and clustering tasks, well-defined quality metrics are available to guide the process. However, such metrics do not exist for regression, making it challenging to split the initial granular regression ball and achieve convergence. 
This highlights the need for a novel metric capable of handling continuous variables to facilitate the effective generation of granular regression balls. 

To address this issue, we propose to first convert the continuous variable to discrete variable in order to replicate the definition of purity considered in the classification scenario. Building on this, we propose a regression-based quality metric for regression. The metric begins by splitting the target variable \( Y \in \mathbb{R} \) into \( k \) nonoverlapping intervals, representing data with similar patterns and therefore giving them the unique label ranging from $1$ to $k$. Thus, the quality of each granular regression ball is defined as:

\begin{equation}
    quality(GRB_j) = \frac{|(GRB_j)|_*}{|GRB_j|}, 
\end{equation}
where, \(|(GRB_j)|_* \) represents the number of samples having majority label in the granular regression ball \( GRB_j \) and $|GRB_j|$ refers to the total number of samples contained in the granular regression ball $GRB_j$, $j=1 \ldots n$. Since the label of granular regression ball is determined by the majority of the samples contained in the GRB, the label will not be affected by the noise or outliers. Therefore, the granular regression ball method is robust.

Using this quality metric, the optimization problem for granular regression ball generation can be formulated as follows:

\begin{equation}
    \quad \underset{n}{min} \left( \sum_{j=1}^n\frac{m}{ |GRB_j|} + n  \right),
\end{equation}
\begin{equation}
    \text{quality}(GRB_j) \geq T, \; j=1,2,\ldots,n
\end{equation}

\noindent 
Here, $n$ is the number of granular regression balls, $T$ is user defined threshold for defining the purity of a ball.

This granular regression ball generation process establishes granular regression  balls as robust summaries of continuous data patterns, paving the way for their integration into regression modeling.

\subsection{Granular Ball Support Vector Regression}

%
Regression samples $D=(X \in \mathbf{R}^l,Y\in \mathbf{R})=\{(x_i,y_i), i=1,2,\dots,m\}$ follows $\epsilon$-insensitive loss function, thus all the points  satisfies the following constraints which depicts the epsilon tube, the upper (lower) tube represented by $l_1$ ( $l_2$) respectively 
\begin{equation}
    l_1: w \cdot x_i + b \ge y_i - \epsilon, i=1,2,\ldots, m,
    \label{eq:svr_cons1}
\end{equation}
\begin{equation}
    l_2: w \cdot x_i + b  \le y_i + \epsilon, i=1,2,\ldots, m .
    \label{eq:svr_cons2}    
\end{equation}

\begin{figure}[h]
\includegraphics[width=0.5\textwidth]{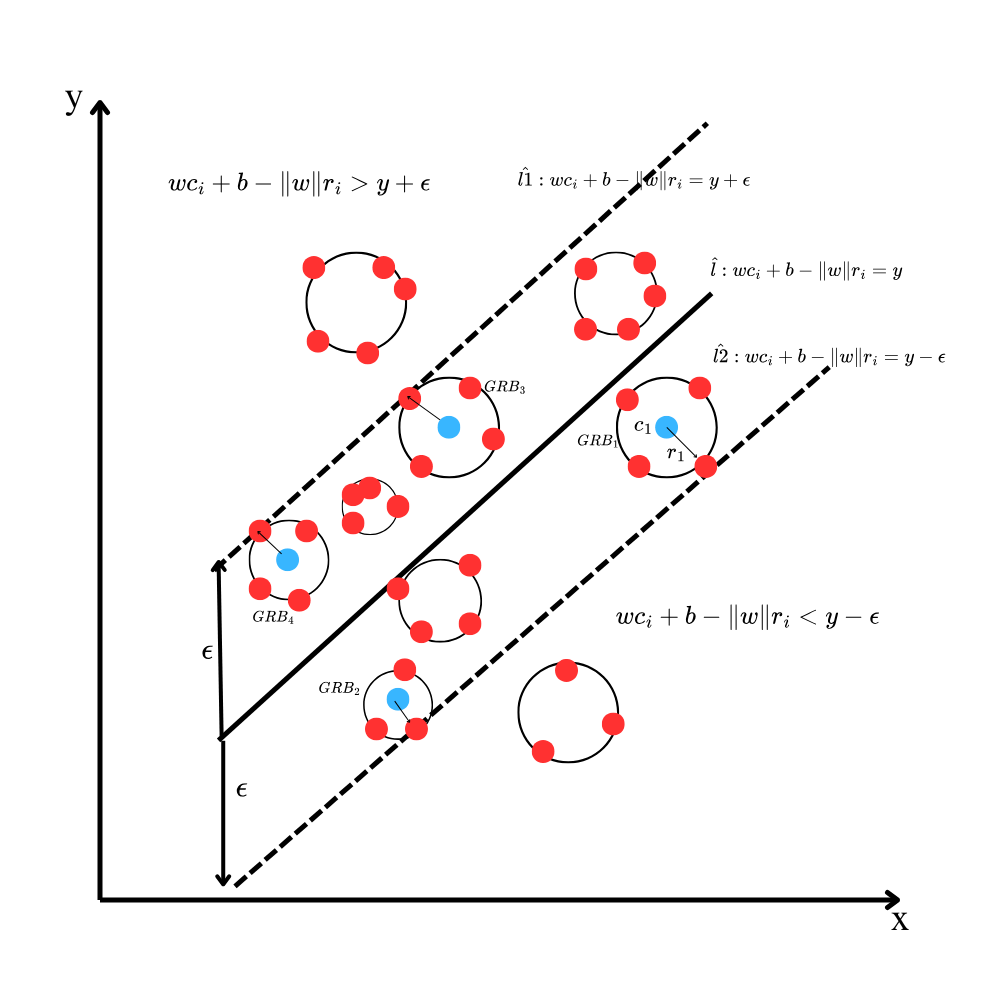}
\caption{ Schematic diagram of the  GBSVR}
\label{gbsvr_wcr}
\end{figure}


Transforming the dataset $D$ into  $n$-granular regression balls ${GRB_i=(c_i,r_i, \hat{y}_i), i=1,2,\ldots,n}, \mbox{such that}~n << m $, where $\hat{y}_i$ is the average of the target points in their respective granular regression ball $GRB_i$.

The distance of the farthest point $x_i$ in the ball $GRB_i$ from the regressor $f$, should be in between $(Y-\epsilon)$ and $(Y + \epsilon)$ as illustrated in Fig. \ref{gbsvr_wcr}. This will guarantee that each ball $GRB_i$ will lie inside the $\epsilon$-tube, as desired. Therefore, the distance of the point $x_i$ from the regressor w.r.t granular regression ball $GRB_i$ is given by 
    


\[
\begin{aligned}
   & \frac{w \cdot (c_i-x_i)}{\|w\|} = r_i,  \\
    & w \cdot c_i - w \cdot x_i = \|w\| \cdot r_i,
\end{aligned}
\]

\begin{equation}
    w \cdot x_i = w \cdot c_i - \|w\| \cdot r_i.
    \label{eq:support_center}
\end{equation}

Substituting  Eq.(\ref{eq:support_center}) in Eq.(\ref{eq:svr_cons1}) and Eq.(\ref{eq:svr_cons2}) leads to the following equations.

\begin{equation}
    \hat{l}_1: \|w\| \cdot r_i + \hat{y}_i - w \cdot c_i - b  \leq \epsilon\;, i=1,2,\ldots, n
    \label{eq:gbsvr_cons1}
\end{equation}
\begin{equation}    
    \hat{l}_2: w \cdot c_i + b -  \hat{y}_i - \|w\| \cdot r_i \leq \epsilon, i=1,2,\ldots, n,
    \label{eq:gbsvr_cons2}    
\end{equation}
here, 
\begin{equation}\label{yhat}
\hat{y}_i=\frac{1}{q}\sum_{j=1}^q y_j, i=1,2,\ldots,n
\end{equation}
$\hat{y}_i$ is the average of $q$ points, $y_j, j=1,2,\ldots, q$, in their respective granular regression balls $GRB_i, i=1,2,\ldots,m$. 


The relationship between the granular regression ball radius \(r\), center \(c\), and the regression plane is derived from the SVR constraints, as shown in Eq.(\ref{eq:gbsvr_cons1}) and Eq.(\ref{eq:gbsvr_cons2}). These constraints define an epsilon-tube, within which the radius \(r\) captures the vertical distance between the bounding points and the regressor.

\subsection{Soft Margin GBSVR}
The objective is to identify regressor such that all support granular regression balls lie outside the $\epsilon$ tube. Since some balls may violate the constraint, soft-margin GBSVR introduces slack variables \( \xi_i \geq 0 \), \( \xi_i^* \geq 0 \) and penalty coefficient $C$ to allow soft errors (violation from some points).
The optimization equation is formulated as:
\begin{eqnarray}
\min_{w, b,\xi_i,\xi_i^*} &\frac{1}{2} \|w\|^2 + C \sum_{i=1}^{n} ( \xi_i + \xi_i^* )
\nonumber\\
\mbox{subject to} &\nonumber\\
&\begin{aligned}
    & \|w\| r_i + \hat{y}_i - w c_i - b  \leq \epsilon + \xi_i,\; i=1,\ldots, n \\
    & w c_i + b -  \hat{y}_i - \|w\| r_i \leq \epsilon + \xi_i^*, \;i=1,\ldots, n \\
    & \xi_i , \xi_i^* \ge 0 
    \label{eq:primal_nonlinear}
\end{aligned}
\end{eqnarray}


Introducing the dual variable $\alpha , \alpha^* , \mu , \mu^* \in \mathbf{R}^n$, the Lagrangian function $\mathcal{L}(w, b, \xi, \xi^*, \alpha, \alpha^*, \mu, \mu^*)$  corresponding to the optimization problem defined in Eq(\ref{eq:primal_nonlinear}) is written as ($\mathcal{L}$ for brevity)

\begin{equation}
\begin{aligned}
\mathcal{L} = & \frac{1}{2} \|w\|^2 + C \sum_{i=1}^{n} (\xi_i + \xi_i^*) \\
& + \sum_{i=1}^{n} \alpha_i \big( \|w\| r_i + \hat{y}_i - w c_i - b - \epsilon - \xi_i \big) \\
& + \sum_{i=1}^{n} \alpha_i^* \big( w c_i + b - \hat{y}_i - \|w\| r_i - \epsilon - \xi_i^* \big) \\
& - \sum_{i=1}^{n} \mu_i \xi_i - \sum_{i=1}^{n} \mu_i^* \xi_i^*.
\label{eq:lagrangian_nonlinear}
\end{aligned}
\end{equation}

The partial derivative of $\mathcal{L}(w, b, \xi, \xi^*, \alpha, \alpha^*, \mu, \mu^*)$ w.r.t primal variables $w , b , \xi , \xi^* $ and equating it with 0 
\begin{align}
\frac{\partial \mathcal{L}}{\partial w} 
&= w - \sum_{i=1}^{n} \alpha_i c_i + \sum_{i=1}^{n} \alpha_i^* c_i = 0,
\label{eq:nonlinear_pd_w} \\
\frac{\partial \mathcal{L}}{\partial b} 
&= -\sum_{i=1}^{n} \alpha_i + \sum_{i=1}^{n} \alpha_i^* = 0,
\label{eq:nonlinear_pd_b} \\
\frac{\partial \mathcal{L}}{\partial \xi_i} 
&= C - \alpha_i - \mu_i = 0, i=1,2,\ldots,n,
\label{eq:nonlinear_pd_xi} \\
\frac{\partial \mathcal{L}}{\partial \xi_i^*} 
&= C - \alpha_i^* - \mu_i^* = 0, i=1,2,\ldots,n.
\label{eq:nonlinear_pd_xi_star}
\end{align}

The Eq.(\ref{eq:nonlinear_pd_w}) can be expressed as 
\begin{equation}
    w = \frac{\|w\| \sum_{i=1}^{n} (\alpha_i - \alpha_i^*)  c_i }{\|w\| + \sum_{i=1}^{n}(\alpha_i - \alpha_i^*) r_i} \label{eq:w_expression}
\end{equation}

To obtain the expression of $w$, we square both sides of Eq.(\ref{eq:w_expression}) leads to:
\begin{equation}
    \|w\|^2 = \frac{\|w\|^2 ( \sum_{i=1}^{n}(\alpha_i - \alpha_i^*)  c_i)^2 }{ ( \|w\| + \sum_{i=1}^{n} (\alpha_i - \alpha_i^*) r_i  )^2 }
\end{equation}

\begin{equation}
   (\sum_{i=1}^{n}(\alpha_i - \alpha_i^*)  c_i)^2  = ( \|w\| + \sum_{i=1}^{n} (\alpha_i - \alpha_i^*) r_i  )^2   \\
    \label{eq:w_equation}
\end{equation}

Taking square root of Eq.(\ref{eq:w_equation}) is :
\begin{equation}
    \| \sum_{i=1}^{n}(\alpha_i - \alpha_i^*)  c_i \| = ( \|w\| + \sum_{i=1}^{n} (\alpha_i - \alpha_i^*) r_i ) \label{eq:reduced_w_eq}
\end{equation}
Since $\|w\| \ge$  0 , $(\alpha_i - \alpha_i^*) \ge$ 0 and $r_i \ge $ 0, Eq.(\ref{eq:reduced_w_eq}) is re-written as :
\begin{equation}
    \|w\| = \| \sum_{i=1}^{n} (\alpha_i - \alpha_i^*) c_i \| - \sum_{i=1}^{n}(\alpha_i - \alpha_i^*) r_i
    \label{eq:normW}
\end{equation}

According to Eq.(\ref{eq:w_expression}) and Eq.(\ref{eq:normW}), on simplification we get, 

   \[ w = \displaystyle{\frac{(\|\sum_{i=1}^{n} (\alpha_i - \alpha_i^*) c_i\| - \sum_{i=1}^{n} (\alpha_i - \alpha_i^*) r_i ) \sum_{i=1}^{n} (\alpha_i - \alpha_i^*) c_i}{\|\sum_{i=1}^{n} (\alpha_i - \alpha_i^*) c_i\|}}\]

\begin{equation}
    w = \frac{(\|A\| - B)A}{\|A\|} \\
     \label{eq:w_value}
\end{equation}

where $A = \sum_{i=1}^{n}(\alpha_i - \alpha_i^*)c_i$ and $B= \sum_{i=1}^{n}(\alpha_i - \alpha_i^*)r_i$ \\

\begin{equation}
    \|w\| = \|A\| - B \\
    \label{eq:value_normW}
\end{equation}




Substituting the value of $w$ in Eq.(\ref{eq:nonlinear_pd_w}), and considering Eq.(\ref{eq:nonlinear_pd_b}), Eq.(\ref{eq:nonlinear_pd_xi}), Eq.(\ref{eq:nonlinear_pd_xi_star}) into Eq.(\ref{eq:lagrangian_nonlinear}), the optimization problem of dual Soft-Margin GBSVR is defined as follows:
\begin{equation}
\begin{aligned}
   & \max_{\alpha, \alpha^*}\   \frac{1}{2} w \cdot w 
    + \sum_{i=1}^{n} (\alpha_i - \alpha_i^*) \|w\| r_i 
    + \sum_{i=1}^{n} (\alpha_i - \alpha_i^*) \hat{y}_i \\
    & + \sum_{i=1}^{n} (C - \alpha_i - \mu_i) + \sum_{i=1}^{n} (C - \alpha_i^* - \mu_i^*) \\
    & - \sum_{i=1}^{n} (\alpha_i - \alpha_i^*) w c_i 
    - \sum_{i=1}^{n} (\alpha_i - \alpha_i^*) b 
    - \sum_{i=1}^{n} (\alpha_i + \alpha_i^*) \epsilon \\
    & = \frac{1}{2} \left(\|A\| - B\right)^2 
    + B (\|A\| - B) 
    + \sum_{i=1}^{n} (\alpha_i - \alpha_i^*) \hat{y}_i \\
    & \quad - \frac{A \left(\|A\| - B\right) A}{\|A\|}
    - \sum_{i=1}^{n} (\alpha_i + \alpha_i^*) \epsilon \\
    & = -\frac{1}{2} \|A\|^2 - \frac{1}{2} B^2 + \|A\| B 
     + \sum_{i=1}^{n} ((\alpha_i - \alpha_i^*) \hat{y}_i 
     -  \epsilon (\alpha_i + \alpha_i^*))
\end{aligned}
\end{equation}
The dual of Eq. (\ref{eq:primal_nonlinear}) is 

\begin{equation}
\begin{aligned}
    \max_{\alpha, \alpha^*} \; & -\frac{1}{2} \|A\|^2 - \frac{1}{2} B^2 
    + \|A\| B + \sum_{i=1}^{n} (\alpha_i - \alpha_i^*) \hat{y}_i \\
    & - \epsilon \sum_{i=1}^{n} (\alpha_i + \alpha_i^*) \\
    \textit{subject to:} \\
    & \ \ \  \sum_{i=1}^{n} (\alpha_i - \alpha_i^*) = 0 \\
    & \ \ \   0 \le \alpha , \alpha^* \le C 
\end{aligned}
\label{eq:dual}
\end{equation}
Using the earlier mathematical derivations, we formalize the optimization steps for granular regression ball generation and GBSVR in Algorithm \ref{algo1} and Algorithm \ref{algo2} respectively.
\subsection{Optimization Algorithm}

\begin{algorithm}[H]
\caption{Granular Regression Ball Generation $GRB_i, i=1,2,\ldots n$} \label{alg:granular_ball}
\begin{algorithmic}[1]
\STATE \textbf{Input:} Data $D=\{X_i, y_i\}_{i=1}^m$, quality threshold $T$, number of labels $k$, minimum number of points in each granular regression ball $p$
\STATE \textbf{Output:} Granular regression balls $GRB_i$, $i=1,2,\ldots n$ with centers $c_i$, radii=$r_i$, and target value $\hat{y}_i, i=1,2, \ldots, n$.
\STATE \textbf{Steps:}
\STATE Split $y_i, i=1,2,\ldots,m$ values into $k$ quantiles and assign new labels $l_i, i=1,2,\ldots,k $ for each quantile
\STATE Initialize a single Granular Regression Ball comprising all the data
\WHILE{there exists a ball that satisfies the splitting criteria}
    \STATE Split the current ball into two smaller balls using KMeans, $K=2$.
    \STATE \textbf{Stopping Criteria:}
    \STATE \hspace{0.5cm} Quality of each ball $\geq T$
    \STATE \hspace{0.5cm} Number of points in each ball $\leq p$
\ENDWHILE
\STATE Return the set of Granular Regression Balls with their centers, radii, and target labels using  (\ref{radius}), (\ref{yhat}), respectively. 
\end{algorithmic}
\label{algo1}
\end{algorithm}

\begin{algorithm}[H]
\caption{Granular Ball Support Vector Regression}\label{alg:gb_svr}
\begin{algorithmic}[1]
\STATE \textbf{Input:} Granular Regression Balls $GRB_i, i=1,2,\ldots n$ with centers, radii, and target values; regularization parameter $C$; tube length $\epsilon$
\STATE \textbf{Output:} Model Weights $w$ and bias term $b$
\STATE \textbf{Steps:}
\STATE Solve for optimization variable $\alpha$, $\alpha^*$  using dual optimization problem (\ref{eq:dual})
\STATE Obtain the dual variables $\alpha$ and $\alpha^*$
\STATE Model parameters are obtained using (\ref{eq:w_value}).
\end{algorithmic}
\label{algo2}
\end{algorithm}

\section{Experiments and Analysis}
All experiments are conducted using Python version 3.12.4 in a Microsoft Windows environment on a machine with a 3.20 GHz CPU and 16 GB RAM. Radial basis function (RBF) kernel is defined as $K(x_1,x_2) = exp^{-\frac{\|x_1 - x_2\|}{2 \sigma^2}}$ where $x_1, x_2 \in \mathbb{R}$ and $\sigma$ is the kernel parameter used for experiments. 
The performance of the proposed algorithm is evaluated and compared against SVR \cite{smola2004tutorial} and NuSVR \cite{williamson2000new} using regression metrics: \( R^2 \), Mean Squared Error(MSE), Mean Absolute Error(MAE), and Root Mean Squared Error(RMSE).
\\












\subsection{Parameter Selection}
For experiments, a grid search was performed to optimize the parameter values. The purity threshold was varied within the set \{0.9, 0.95, 0.97, 0.99, 0.995, 0.997\}, while the minimum number of points in a ball was tested with values of $2$, $3$, and $4$. The $\epsilon$ in $\epsilon$-tube was selected from the set \(\{10^{-i} \mid i = 1, 2, 3, \ldots, 9\}\). The value of the kernel parameter was chosen from the interval \( [10^{-3}: 1] \) with the step size as $0.01$. 

\subsection{Synthetic Datasets}
The $sinc$ function and the $cos$ function with noise are frequently used for synthetic data generation to evaluate the performance of a regression model and is defined as
\begin{align}
    \text{(Type A)} \quad y_i = \frac{\sin(\pi x_i)}{\pi x_i} + \eta, \quad x_i \sim U[-4, 4]\\
    \text{(Type B)} \quad y_i = \cos(\pi x_i) + \eta, \quad x_i \sim U[-1, 1]
\end{align}

To verify the GBSVR's effectiveness, six types of heteroscedastic noise are applied, as follows:

\[
\begin{aligned}
\text{Type 1: } \eta_i &= \left(-\frac{|x_i|}{8} + 0.5\right) e_i, \, e_i \sim N(0, 0.15^2); \\
\text{Type 2: } \eta_i &= \left(-\frac{|x_i|}{8} + 0.5\right) e_i, \, e_i \sim U[-0.25, 0.25]; \\
\text{Type 3: } \eta_i &= \left(-\frac{|x_i|}{8} + 0.5\right) e_i, \, e_i \sim N(0, 0.02^2); \\
\text{Type 4: } \eta_i &= \left(-\frac{|x_i|}{8} + 0.5\right) e_i, \, e_i \sim U[-0.02, 0.02]; \\
\text{Type 5: } \eta_i &= \left(-\frac{|x_i|}{8} + 0.5\right) e_i, \, e_i \sim N(0, 0.12^2); \\
\text{Type 6: } \eta_i &= \left(-\frac{|x_i|}{8} + 0.5\right) e_i, \, e_i \sim U[-0.2, 0.2].
\end{aligned}
\]
where \(N(c, d^2)\) is defined as the Gaussian random variable with mean \(c\) and variance \(d^2\), and \(U[m, n]\) is the uniform random variable defined on the interval \([m, n]\).

\begin{table}[ht]
\scriptsize
\centering
\caption{Comparison results of different models with six different types of noise  on Type A dataset.}
\renewcommand{\arraystretch}{1.2}
\setlength{\tabcolsep}{4pt}
\begin{tabular}{|l|cccc|cccc|}
\hline
\multirow{2}{*}{\textbf{Model}} & \multicolumn{4}{c|}{\textbf{Type 1 Noise}} & \multicolumn{4}{c|}{\textbf{Type 2 Noise}} \\
\cline{2-9}
& \textbf{R$^2$} & \textbf{MAE} & \textbf{MSE} & \textbf{RMSE} & \textbf{R$^2$} & \textbf{MAE} & \textbf{MSE} & \textbf{RMSE} \\
\hline
GBSVR & \textbf{0.9823} & \textbf{0.0345} & \textbf{0.0021} & \textbf{0.0459} & \textbf{0.9776} & \textbf{0.0389} & \textbf{0.0025} & \textbf{0.0503} \\
SVR & 0.9760 & 0.0427 & 0.0028 & 0.0528 & 0.9662 & 0.0462 & 0.0038 & 0.0613 \\
NuSVR & 0.9769 & 0.0403 & 0.0027 & 0.0522 & 0.9746 & 0.0414 & 0.0027 & 0.0519 \\
\hline

\hline
\multirow{2}{*}{\textbf{Model}} & \multicolumn{4}{c|}{\textbf{Type 3 Noise}} & \multicolumn{4}{c|}{\textbf{Type 4 Noise}} \\
\cline{2-9}
& \textbf{R$^2$} & \textbf{MAE} & \textbf{MSE} & \textbf{RMSE} & \textbf{R$^2$} & \textbf{MAE} & \textbf{MSE} & \textbf{RMSE} \\
\hline
GBSVR & \textbf{0.9994} & \textbf{0.0066} & \textbf{0.0001} & \textbf{0.0081} & \textbf{0.9995} & \textbf{0.0058} & \textbf{0.0001} & \textbf{0.0074} \\
SVR & 0.9972 & 0.0141 & 0.0003 & 0.0177 & 0.9967 & 0.0158 & 0.0004 & 0.0195 \\
NuSVR & 0.9935 & 0.0222 & 0.0007 & 0.0261 & 0.9997 & 0.0048 & 0.0000 & 0.0058 \\
\hline

\hline
\multirow{2}{*}{\textbf{Model}} & \multicolumn{4}{c|}{\textbf{Type 5 Noise}} & \multicolumn{4}{c|}{\textbf{Type 6 Noise}} \\
\cline{2-9}
& \textbf{R$^2$} & \textbf{MAE} & \textbf{MSE} & \textbf{RMSE} & \textbf{R$^2$} & \textbf{MAE} & \textbf{MSE} & \textbf{RMSE} \\
\hline
GBSVR & \textbf{0.9884} & \textbf{0.0276} & \textbf{0.0014} & \textbf{0.0369} & 0.9995 & 0.0058 & 0.0001 & 0.0074 \\
SVR & 0.9829 & 0.0343 & 0.0020 & 0.0444 & 0.9967 & 0.0158 & 0.0004 & 0.0195 \\
NuSVR & 0.9540 & 0.0613 & 0.0057 & 0.0755 & \textbf{0.9997} & \textbf{0.0048} & \textbf{0.0000} & \textbf{0.0058} \\
\hline
\end{tabular}
\label{sinc_table}
\end{table}

\begin{table}[ht]
\scriptsize
\centering
\caption{Comparison results of three models with six different types of noise on Type B dataset.}
\renewcommand{\arraystretch}{1.2}
\setlength{\tabcolsep}{4pt}
\begin{tabular}{|l|cccc|cccc|}
\hline
\multirow{2}{*}{\textbf{Model}} & \multicolumn{4}{c|}{\textbf{Type 1 Noise}} & \multicolumn{4}{c|}{\textbf{Type 2 Noise}} \\
\cline{2-9}
& \textbf{R$^2$} & \textbf{MAE} & \textbf{MSE} & \textbf{RMSE} & \textbf{R$^2$} & \textbf{MAE} & \textbf{MSE} & \textbf{RMSE} \\
\hline
GBSVR & \textbf{0.9919} & \textbf{0.0523} & \textbf{0.0041} & \textbf{0.0641} & \textbf{0.9920} & \textbf{0.0527} & \textbf{0.0039} & \textbf{0.0624} \\
SVR   & 0.9867 & 0.0682 & 0.0067 & 0.0816 & 0.9868 & 0.0651 & 0.0065 & 0.0805 \\
NuSVR & 0.9653 & 0.1181 & 0.0183 & 0.1353 & 0.9756 & 0.0946 & 0.0127 & 0.1125 \\
\hline

\hline
\multirow{2}{*}{\textbf{Model}} & \multicolumn{4}{c|}{\textbf{Type 3 Noise}} & \multicolumn{4}{c|}{\textbf{Type 4 Noise}} \\
\cline{2-9}
& \textbf{R$^2$} & \textbf{MAE} & \textbf{MSE} & \textbf{RMSE} & \textbf{R$^2$} & \textbf{MAE} & \textbf{MSE} & \textbf{RMSE} \\
\hline
GBSVR & \textbf{0.9998} & \textbf{0.0085} & \textbf{0.0001} & \textbf{0.0103} & \textbf{0.9999} & 0.0059 & 0.0001 & 0.0072 \\
SVR   & 0.9996 & 0.0112 & 0.0002 & 0.0133 & 0.9999 & \textbf{0.0052} & \textbf{0.0000} & \textbf{0.0066} \\
NuSVR & 0.9996 & 0.0114 & 0.0002 & 0.0137 & 0.9999 & 0.0058 & 0.0001 & 0.0071 \\
\hline

\hline
\multirow{2}{*}{\textbf{Model}} & \multicolumn{4}{c|}{\textbf{Type 5 Noise}} & \multicolumn{4}{c|}{\textbf{Type 6 Noise}} \\
\cline{2-9}
& \textbf{R$^2$} & \textbf{MAE} & \textbf{MSE} & \textbf{RMSE} & \textbf{R$^2$} & \textbf{MAE} & \textbf{MSE} & \textbf{RMSE} \\
\hline
GBSVR & \textbf{0.9945} & \textbf{0.0429} & \textbf{0.0028} & \textbf{0.0525} & \textbf{0.9939} & \textbf{0.0444} & \textbf{0.0029} & \textbf{0.0535} \\
SVR   & 0.9916 & 0.0541 & 0.0043 & 0.0657 & 0.9918 & 0.0516 & 0.0041 & 0.0644 \\
NuSVR & 0.9779 & 0.0912 & 0.0107 & 0.1035 & 0.9812 & 0.0799 & 0.0091 & 0.0952 \\
\hline
\end{tabular}
\label{cos_table}
\end{table}

    Table \ref{sinc_table} and Table \ref{cos_table} present the results for Type A and Type B data under various noise conditions. The performance of GBSVR demonstrates its robustness, consistently outperforming or matching over other methods.

 
\subsection{UCI benchmark dataset}
To illustrate the effectiveness of the proposed methodology in various domains and applications, we performed regression experiments on a variety of benchmark datasets. These datasets include UCI datasets \cite{UCI_ML_Repository}, which are widely utilized to evaluate algorithms' efficiency, are mentioned in Table \ref{tab:dataset_info}.

The results were obtained using a $5$-fold cross-validation strategy, employing the RBF kernel for GBSVR, SVR, and NuSVR. The outcomes, presented in Table \ref{tab:comparison} , highlight the performance of GBSVR, SVR, and NuSVR on various datasets across regression metrics: \(R^2\), MSE, MAE, and RMSE. The findings demonstrate that GBSVR consistently outperformed the other algorithms in all metrics in  all UCI datasets. This highlights the effectiveness of granular regression ball methodology in generalizing data by representing a cluster of data samples as a single ball rather than relying on individual data samples. The table also illustrates the impact of adding Gaussian noise, with a mean of $0$ and a standard deviation of $0.2$, to a subset of the training samples and compares the performance of the algorithms under these conditions. The percentage of noisy samples is incrementally increased from $0\%$ to $20\%$. The proposed algorithm shows minimal performance degradation as the proportion of noisy samples increases, while the performance of SVR diminishes significantly with higher levels of noise. This demonstrates the ability of GBSVR to effectively handle noisy data.

\begin{table}[h]
    \centering
    
    \caption{Dataset Information}
    \begin{tabular}{lcc}
        \toprule
        \textbf{Data Set} & \textbf{Dimensions} & \textbf{Number of Balls} \\
        \midrule
        Real Estate Valuation & 414 x 6 & 147 \\
        AutoMPG & 392 x 7 & 137 \\
        Autos & 159 x 25 & 59 \\
        Servo & 167 x 4 & 56 \\
        Yacht & 308 x 6 & 56 \\
        Machine & 209 x 7 & 74 \\
        \bottomrule
    \end{tabular}
    \label{tab:dataset_info}
\end{table}

Table \ref{tab:comparison}  also reports the CPU time required by each algorithm. It can be observed that the time taken by GBSVR is on average almost ten times less than that of SVR and NuSVR, a trend consistent across all UCI datasets. This performance improvement is attributed to the granular regression ball representation employed by GBSVR, which significantly reduces its overall time complexity. In contrast, SVR and NuSVR process each data point individually, resulting in higher computational time requirements.

\hspace{-1em}

\begin{table*}[]
\scriptsize
\centering
\caption{Comparison of Performance Between Different Methods with Different Levels of Noise}
\renewcommand{\arraystretch}{1.3}
\setlength{\tabcolsep}{3pt}
\resizebox{1.05\textwidth}{!}{%

\begin{tabular}{|c|c|ccccc|c|c|ccccc|}
\hline

\multicolumn{7}{|c|}{\textbf{Servo}} & \multicolumn{7}{c|}{\textbf{Yacht}} \\
\hline
\multicolumn{2}{|c|}{} & \multicolumn{5}{c|}{\textbf{Noise Percentage}} & \multicolumn{2}{c|}{} & \multicolumn{5}{c|}{\textbf{Noise Percentage}} \\
\hline
\textbf{Metric} & \textbf{Methods} & 0 & 0.05 & 0.1 & 0.15 & 0.2 & \textbf{Metric} & \textbf{Methods} & 0 & 0.05 & 0.1 & 0.15 & 0.2 \\
\hline

\multirow{3}{*}{\textbf{Time $(\downarrow)$}} 
& \textbf{GBSVR} & \textbf{0.996±0.058} & \textbf{1.039±0.069} & \textbf{1.023±0.063} & \textbf{1.088±0.08} & \textbf{1.082±0.049} 
& \multirow{3}{*}{\textbf{Time $(\downarrow)$}} 
& \textbf{GBSVR} & \textbf{1.194±0.29} & \textbf{1.076±0.274} & \textbf{1.003±0.27} & \textbf{1.102±0.334} & \textbf{1.187±0.283} \\
& \textbf{SVR} & 9.289±0.116 & 9.3±0.098 & 9.319±0.124 & 9.374±0.097 & 9.351±0.095 
& & \textbf{SVR} & 41.285±0.356 & 41.118±0.441 & 40.199±0.453 & 42.357±1.694 & 41.433±0.395 \\
& \textbf{NuSVR} & 4.729±0.104 & 4.902±0.231 & 4.903±0.23 & 4.889±0.291 & 4.612±0.281 
& & \textbf{NuSVR} & 37.262±0.179 & 37.798±0.23 & 37.588±0.425 & 40.393±1.299 & 37.534±0.235 \\
\hline

\multirow{3}{*}{\textbf{$R^2$ $(\uparrow)$}} 
& \textbf{GBSVR} & \textbf{0.855±0.014} & \textbf{0.856±0.032} & \textbf{0.863±0.031} & \textbf{0.834±0.032} & \textbf{0.846±0.039} 
& \multirow{3}{*}{\textbf{$R^2$ $(\uparrow)$}} 
& \textbf{GBSVR} & \textbf{0.968±0.014} & \textbf{0.969±0.015} & \textbf{0.968±0.015} & \textbf{0.966±0.015} & \textbf{0.968±0.014} \\
& \textbf{SVR} & 0.773±0.08 & 0.707±0.124 & 0.739±0.115 & 0.612±0.199 & 0.756±0.048 
& & \textbf{SVR} & 0.962±0.021 & 0.966±0.017 & 0.952±0.016 & 0.960±0.008 & 0.962±0.021 \\
& \textbf{NuSVR} & 0.836±0.03 & 0.833±0.03 & 0.817±0.052 & 0.811±0.041 & 0.833±0.044 
& & \textbf{NuSVR} & 0.966±0.016 & 0.964±0.015 & 0.963±0.017 & 0.964±0.015 & 0.966±0.016 \\
\hline

\multirow{3}{*}{\textbf{MAE $(\downarrow)$}} 
& \textbf{GBSVR} & \textbf{0.274±0.009} & \textbf{0.262±0.024} & \textbf{0.256±0.038} & \textbf{0.286±0.023} & \textbf{0.276±0.036} 
& \multirow{3}{*}{\textbf{MAE $(\downarrow)$}} 
& \textbf{GBSVR} & \textbf{0.105±0.021} & \textbf{0.104±0.019} & \textbf{0.106±0.018} & \textbf{0.111±0.02} & \textbf{0.105±0.021} \\
& \textbf{SVR} & 0.366±0.063 & 0.424±0.104 & 0.384±0.047 & 0.473±0.124 & 0.394±0.077 
& & \textbf{SVR} & 0.13±0.031 & 0.122±0.023 & 0.16±0.035 & 0.143±0.009 & 0.13±0.031 \\
& \textbf{NuSVR} & 0.298±0.019 & 0.301±0.028 & 0.325±0.049 & 0.323±0.021 & 0.305±0.033 
& & \textbf{NuSVR} & 0.105±0.02 & 0.107±0.016 & 0.111±0.015 & 0.112±0.015 & 0.105±0.02 \\
\hline

\multirow{3}{*}{\textbf{RMSE $(\downarrow)$}} 
& \textbf{GBSVR} & \textbf{0.376±0.029} & \textbf{0.370±0.032} & \textbf{0.365±0.062} & \textbf{0.399±0.040} & \textbf{0.385±0.062} 
& \multirow{3}{*}{\textbf{RMSE $(\downarrow)$}} 
& \textbf{GBSVR} & \textbf{0.173±0.051} & \textbf{0.17±0.051} & \textbf{0.172±0.052} & \textbf{0.178±0.05} & \textbf{0.173±0.051} \\
& \textbf{SVR} & 0.459±0.077 & 0.524±0.120 & 0.487±0.067 & 0.599±0.136 & 0.488±0.073 
& & \textbf{SVR} & 0.187±0.064 & 0.176±0.054 & 0.214±0.051 & 0.199±0.033 & 0.187±0.064 \\
& \textbf{NuSVR} & 0.397±0.038 & 0.402±0.042 & 0.418±0.067 & 0.425±0.036 & 0.400±0.059 
& & \textbf{NuSVR} & 0.179±0.052 & 0.185±0.049 & 0.186±0.052 & 0.184±0.046 & 0.179±0.052 \\
\hline

\multicolumn{7}{|c|}{\textbf{Autompg}} & \multicolumn{7}{c|}{\textbf{Autos}} \\
\hline
\multicolumn{2}{|c|}{} & \multicolumn{5}{c|}{\textbf{Noise Percentage}} & \multicolumn{2}{c|}{} & \multicolumn{5}{c|}{\textbf{Noise Percentage}} \\
\hline
\textbf{Metric} & \textbf{Methods} & 0 & 0.05 & 0.1 & 0.15 & 0.2 & \textbf{Metric} & \textbf{Methods} & 0 & 0.05 & 0.1 & 0.15 & 0.2 \\
\hline

\multirow{3}{*}{\textbf{Time $(\downarrow)$}} 
& \textbf{GBSVR} & \textbf{7.234±0.524} & \textbf{6.813±0.502} & \textbf{7.46±0.511} & \textbf{7.664±0.587} & \textbf{7.029±0.556} 
& \multirow{3}{*}{\textbf{Time $(\downarrow)$}} 
& \textbf{GBSVR} & \textbf{1.11±0.112} & \textbf{1.113±0.152} & \textbf{1.111±0.115} & \textbf{1.11±0.174} & \textbf{1.227±0.139} \\
& \textbf{SVR} & 69.362±0.425 & 69.68±0.464 & 70.256±0.515 & 70.323±0.466 & 71.437±0.469 
& & \textbf{SVR} & 9.416±0.371 & 9.389±0.285 & 9.124±0.354 & 9.468±0.568 & 9.576±0.456 \\
& \textbf{NuSVR} & 85.806±0.699 & 85.985±0.661 & 45.173±4.906 & 34.241±2.73 & 31.724±3.354 
& & \textbf{NuSVR} & 8.464±0.559 & 8.401±0.45 & 8.276±0.765 & 8.438±0.589 & 8.564±0.549 \\
\hline

\multirow{3}{*}{\textbf{$R^2$ $(\uparrow)$}} 
& \textbf{GBSVR} & \textbf{0.866±0.016} & \textbf{0.848±0.019} & \textbf{0.864±0.012} & \textbf{0.846±0.028} & \textbf{0.84±0.023} 
& \multirow{3}{*}{\textbf{$R^2$ $(\uparrow)$}} 
& \textbf{GBSVR} & \textbf{0.872±0.037} & \textbf{0.873±0.039} & \textbf{0.872±0.025} & \textbf{0.89±0.024} & \textbf{0.875±0.028} \\
& \textbf{SVR} & 0.787±0.056 & 0.777±0.037 & 0.663±0.297 & 0.754±0.05 & 0.753±0.058 
& & \textbf{SVR} & 0.74±0.068 & 0.725±0.093 & 0.701±0.087 & 0.789±0.068 & 0.715±0.137 \\
& \textbf{NuSVR} & 0.812±0.028 & 0.81±0.028 & 0.817±0.033 & 0.817±0.031 & 0.802±0.034 
& & \textbf{NuSVR} & 0.839±0.035 & 0.845±0.033 & 0.829±0.047 & 0.824±0.05 & 0.835±0.052 \\
\hline

\multirow{3}{*}{\textbf{MAE $(\downarrow)$}} 
& \textbf{GBSVR} & \textbf{0.27±0.028} & \textbf{0.276±0.021} & \textbf{0.268±0.022} & \textbf{0.283±0.046} & \textbf{0.298±0.035} 
& \multirow{3}{*}{\textbf{MAE $(\downarrow)$}} 
& \textbf{GBSVR} & \textbf{0.265±0.032} & \textbf{0.264±0.017} & \textbf{0.266±0.034} & \textbf{0.253±0.026} & \textbf{0.262±0.021} \\
& \textbf{SVR} & 0.364±0.068 & 0.367±0.018 & 0.428±0.187 & 0.366±0.038 & 0.379±0.052 
& & \textbf{SVR} & 0.356±0.033 & 0.353±0.033 & 0.367±0.035 & 0.322±0.034 & 0.355±0.056 \\
& \textbf{NuSVR} & 0.347±0.029 & 0.35±0.028 & 0.311±0.03 & 0.309±0.047 & 0.311±0.033 
& & \textbf{NuSVR} & 0.285±0.039 & 0.286±0.046 & 0.292±0.042 & 0.301±0.041 & 0.284±0.046 \\
\hline

\multirow{3}{*}{\textbf{RMSE $(\downarrow)$}} 
& \textbf{GBSVR} & \textbf{0.363±0.039} & \textbf{0.386±0.028} & \textbf{0.365±0.023} & \textbf{0.389±0.049} & \textbf{0.397±0.049} 
& \multirow{3}{*}{\textbf{RMSE $(\downarrow)$}} 
& \textbf{GBSVR} & \textbf{0.344±0.032} & \textbf{0.339±0.023} & \textbf{0.347±0.036} & \textbf{0.321±0.031} & \textbf{0.341±0.021} \\
& \textbf{SVR} & 0.455±0.082 & 0.464±0.024 & 0.533±0.207 & 0.488±0.041 & 0.489±0.057 
& & \textbf{SVR} & 0.493±0.076 & 0.5±0.079 & 0.521±0.059 & 0.436±0.037 & 0.499±0.103 \\
& \textbf{NuSVR} & 0.429±0.041 & 0.432±0.04 & 0.424±0.054 & 0.425±0.063 & 0.441±0.061 
& & \textbf{NuSVR} & 0.391±0.065 & 0.384±0.068 & 0.397±0.056 & 0.404±0.064 & 0.394±0.086 \\
\hline

\multicolumn{7}{|c|}{\textbf{Machine}} & \multicolumn{7}{c|}{\textbf{Real Estate Valuation}} \\
\hline
\multicolumn{2}{|c|}{} & \multicolumn{5}{c|}{\textbf{Noise Percentage}} & \multicolumn{2}{c|}{} & \multicolumn{5}{c|}{\textbf{Noise Percentage}} \\
\hline
\textbf{Metric} & \textbf{Methods} & 0 & 0.05 & 0.1 & 0.15 & 0.2 & \textbf{Metric} & \textbf{Methods} & 0 & 0.05 & 0.1 & 0.15 & 0.2 \\
\hline

\multirow{3}{*}{\textbf{Time $(\downarrow)$}} 
& \textbf{GBSVR} & \textbf{1.888±0.115} & \textbf{1.902±0.136} & \textbf{1.918±0.126} & \textbf{1.993±0.177} & \textbf{2.069±0.153} 
& \multirow{3}{*}{\textbf{Time $(\downarrow)$}} 
& \textbf{GBSVR} & \textbf{7.335±1.151} & \textbf{7.121±0.633} & \textbf{6.86±0.881} & \textbf{8.004±1.186} & \textbf{7.522±0.642} \\
& \textbf{SVR} & 17.078±0.233 & 16.954±0.226 & 16.993±0.262 & 17.053±0.229 & 16.952±0.251 
& & \textbf{SVR} & 95.774±0.917 & 94.793±0.865 & 94.927±0.703 & 96.663±1.722 & 95.103±0.671 \\
& \textbf{NuSVR} & 17.222±0.269 & 17.12±0.297 & 17.165±0.261 & 17.2±0.248 & 17.13±0.301 
& & \textbf{NuSVR} & 93.068±1.192 & 92.017±1.501 & 91.805±0.768 & 94.193±1.282 & 91.832±0.746 \\
\hline

\multirow{3}{*}{\textbf{$R^2$ $(\uparrow)$}} 
& \textbf{GBSVR} & \textbf{0.836±0.021} & \textbf{0.834±0.027} & \textbf{0.835±0.032} & \textbf{0.82±0.042} & \textbf{0.826±0.036} 
& \multirow{3}{*}{\textbf{$R^2$ $(\uparrow)$}} 
& \textbf{GBSVR} & \textbf{0.63±0.075} & \textbf{0.629±0.069} & \textbf{0.633±0.077} & \textbf{0.637±0.073} & \textbf{0.625±0.071} \\
& \textbf{SVR} & 0.744±0.065 & 0.743±0.053 & 0.718±0.103 & 0.734±0.057 & 0.746±0.046 
& & \textbf{SVR} & 0.542±0.059 & 0.546±0.07 & 0.535±0.062 & 0.555±0.066 & 0.531±0.057 \\
& \textbf{NuSVR} & 0.746±0.035 & 0.748±0.036 & 0.753±0.037 & 0.749±0.036 & 0.747±0.034 
& & \textbf{NuSVR} & 0.548±0.061 & 0.557±0.071 & 0.54±0.065 & 0.553±0.059 & 0.53±0.06 \\
\hline

\multirow{3}{*}{\textbf{MAE $(\downarrow)$}} 
& \textbf{GBSVR} & \textbf{0.311±0.023} & \textbf{0.311±0.03} & \textbf{0.308±0.023} & \textbf{0.321±0.023} & \textbf{0.32±0.021} 
& \multirow{3}{*}{\textbf{MAE $(\downarrow)$}} 
& \textbf{GBSVR} & \textbf{0.403±0.034} & \textbf{0.4±0.032} & \textbf{0.4±0.033} & \textbf{0.395±0.032} & \textbf{0.403±0.03} \\
& \textbf{SVR} & 0.376±0.071 & 0.384±0.068 & 0.388±0.078 & 0.385±0.069 & 0.379±0.062 
& & \textbf{SVR} & 0.503±0.035 & 0.5±0.037 & 0.507±0.035 & 0.499±0.039 & 0.508±0.03 \\
& \textbf{NuSVR} & 0.392±0.033 & 0.392±0.033 & 0.39±0.037 & 0.392±0.036 & 0.389±0.036 
& & \textbf{NuSVR} & 0.5±0.034 & 0.495±0.04 & 0.504±0.038 & 0.498±0.03 & 0.511±0.036 \\
\hline

\multirow{3}{*}{\textbf{RMSE $(\downarrow)$}} 
& \textbf{GBSVR} & \textbf{0.4±0.033} & \textbf{0.401±0.031} & \textbf{0.399±0.033} & \textbf{0.414±0.027} & \textbf{0.409±0.031} 
& \multirow{3}{*}{\textbf{RMSE $(\downarrow)$}} 
& \textbf{GBSVR} & \textbf{0.604±0.099} & \textbf{0.605±0.094} & \textbf{0.601±0.1} & \textbf{0.598±0.096} & \textbf{0.608±0.096} \\
& \textbf{SVR} & 0.499±0.091 & 0.501±0.079 & 0.522±0.126 & 0.51±0.082 & 0.498±0.067 
& & \textbf{SVR} & 0.671±0.069 & 0.667±0.071 & 0.676±0.073 & 0.661±0.075 & 0.679±0.072 \\
& \textbf{NuSVR} & 0.496±0.031 & 0.494±0.03 & 0.489±0.036 & 0.493±0.033 & 0.495±0.033 
& & \textbf{NuSVR} & 0.666±0.069 & 0.659±0.075 & 0.672±0.075 & 0.663±0.069 & 0.68±0.07 \\
\hline

\end{tabular}
}
\label{tab:comparison}
\end{table*}

The granular regression ball methodology contributes to the improved performance of GBSVR by improving generalization and robustness, especially in noisy environments. By consolidating data points into clusters, GBSVR reduces sensitivity to noise and outliers, making it more resilient to perturbations in the data. Furthermore, this aggregation reduces computational time, allowing GBSVR to handle large datasets more efficiently compared to other methods.

\subsection{Stock Price Prediction}
To demonstrate GBSVR's effectiveness, it is applied to stock price prediction, a highly volatile and nonlinear task with noisy financial data. The dataset comprises five years of historical stock prices (Jan 2019–Jan 2025), with the Closing Price as the target. A sliding window approach (window length = $5$) generates features, where the first five values form the input, and the sixth is the target. The model predicts the $7$th value in the sequence. Training uses $30\%$ of the data, while $70\%$ is reserved for testing.

\begin{figure}[H]
    \centering
    \begin{minipage}{0.35\textwidth}
        \centering
        \includegraphics[width=\linewidth]{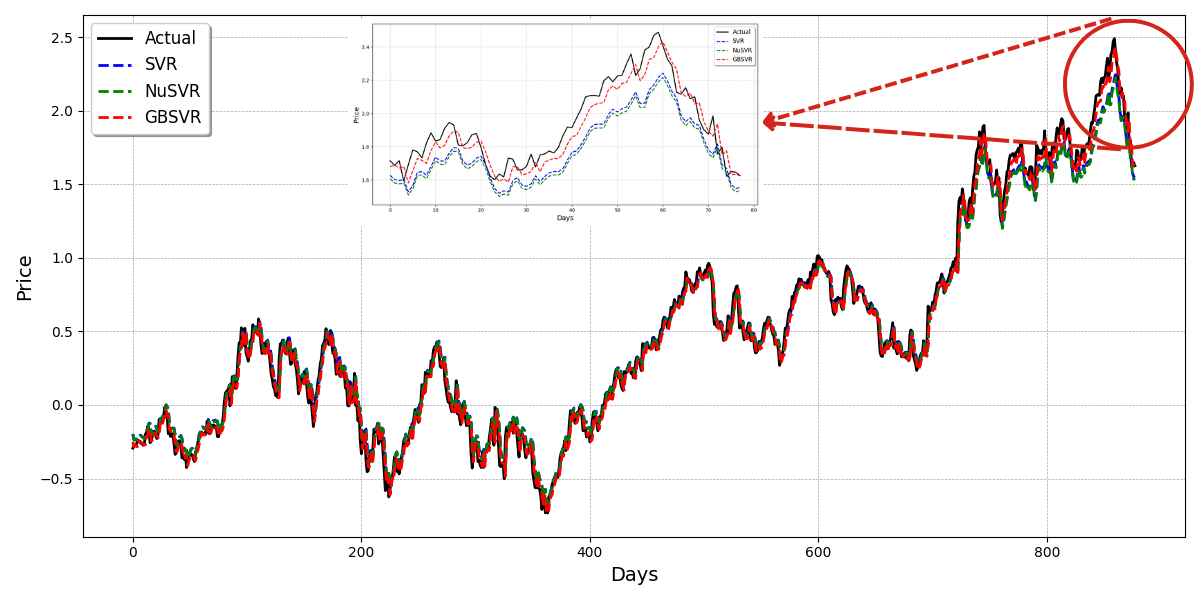}
        \text{APPLE}
    \end{minipage}
    \hfill
    \begin{minipage}{0.35\textwidth}
        \centering
        \includegraphics[width=\linewidth]{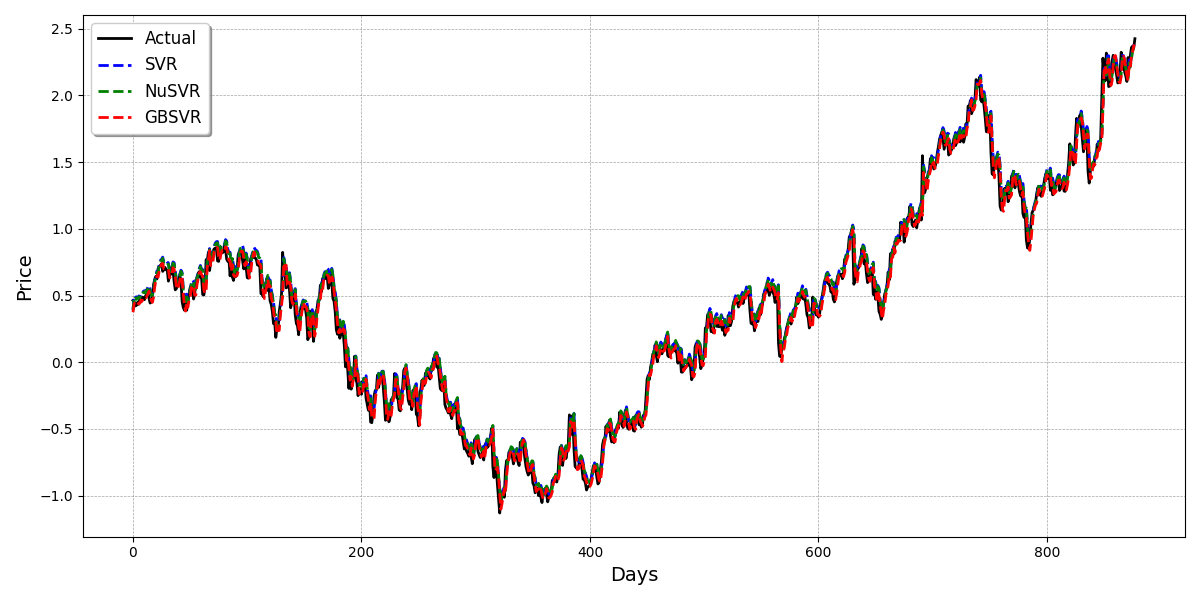}
        \text{GOOGLE}
    \end{minipage}
    
    \caption{Stock Price Forecasting (APPLE and GOOGLE)}
    \label{stock_price_1}
\end{figure}

Table \ref{stocks_table} compares the performance of GBSVR, SVR, and NuSVR across different stocks. GBSVR outperforms the other models on all metrics for every stock. Fig. \ref{stock_price_1} shows actual and predicted values from the three models over $878$ days.

\begin{table}[H]
\scriptsize
\centering
\caption{Comparison results of three models for stock forecasting on Financial datasets.}
\renewcommand{\arraystretch}{1.2}
\setlength{\tabcolsep}{4pt}
\begin{tabular}{|l|cccc|cccc|}
\hline
\multirow{2}{*}{\textbf{Model}} & \multicolumn{4}{c|}{\textbf{Apple}} & \multicolumn{4}{c|}{\textbf{Google}} \\
\cline{2-9}
& \textbf{R$^2$} & \textbf{MAE} & \textbf{MSE} & \textbf{RMSE} & \textbf{R$^2$} & \textbf{MAE} & \textbf{MSE} & \textbf{RMSE} \\
\hline
GBSVR & \textbf{0.9814} & \textbf{0.0623} & \textbf{0.0072} & \textbf{0.0847} & \textbf{0.9272} & \textbf{0.0864} & \textbf{0.0167} & \textbf{0.1292} \\
SVR   & 0.9782 & 0.0710 & 0.0084 & 0.0917 & 0.8825 & 0.1165 & 0.0269 & 0.1641 \\
NuSVR & 0.9784 & 0.0737 & 0.0083 & 0.0913 & 0.8960 & 0.1237 & 0.0238 & 0.1544 \\
\hline
\end{tabular}
\vspace{0.5cm}

\begin{tabular}{|l|cccc|cccc|}
\hline
\multirow{2}{*}{\textbf{Model}} & \multicolumn{4}{c|}{\textbf{NVIDIA}} & \multicolumn{4}{c|}{\textbf{Tesla}} \\
\cline{2-9}
& \textbf{R$^2$} & \textbf{MAE} & \textbf{MSE} & \textbf{RMSE} & \textbf{R$^2$} & \textbf{MAE} & \textbf{MSE} & \textbf{RMSE} \\
\hline
GBSVR & \textbf{0.9661} & \textbf{0.0900} & \textbf{0.0142} & \textbf{0.1193} & \textbf{0.9727} & \textbf{0.1074} & \textbf{0.0238} & \textbf{0.1543} \\
SVR   & 0.9434 & 0.1280 & 0.0238 & 0.1542 & 0.9678 & 0.1340 & 0.0281 & 0.1676 \\
NuSVR & 0.9468 & 0.1238 & 0.0223 & 0.1494 & 0.9680 & 0.1362 & 0.0280 & 0.1672 \\
\hline
\end{tabular}
\label{stocks_table}
\end{table}

\subsection{Short-term wind speed prediction using real world dataset}
To predict wind speed, the dataset utilized consists of 36,000 samples collected over a span of $25$ days, with measurements taken every minute. Of this dataset, $80\%$ (28,800 samples) was used for training, while the remaining $20\%$ ($7,200$ samples) was allocated for testing.

The wind speed prediction model was constructed using the following approach, reflecting the actual scenario: the input vector at time step \(i\) is defined as $\vec{x_i} = (x_{i-4}, x_{i-3}, x_{i-2}, x_{i-1}), \quad i = 5, \ldots, 36000$.

\begin{figure}[ht]
    \centering
    \begin{minipage}{0.45\textwidth}
        \centering
        \includegraphics[width=\linewidth]{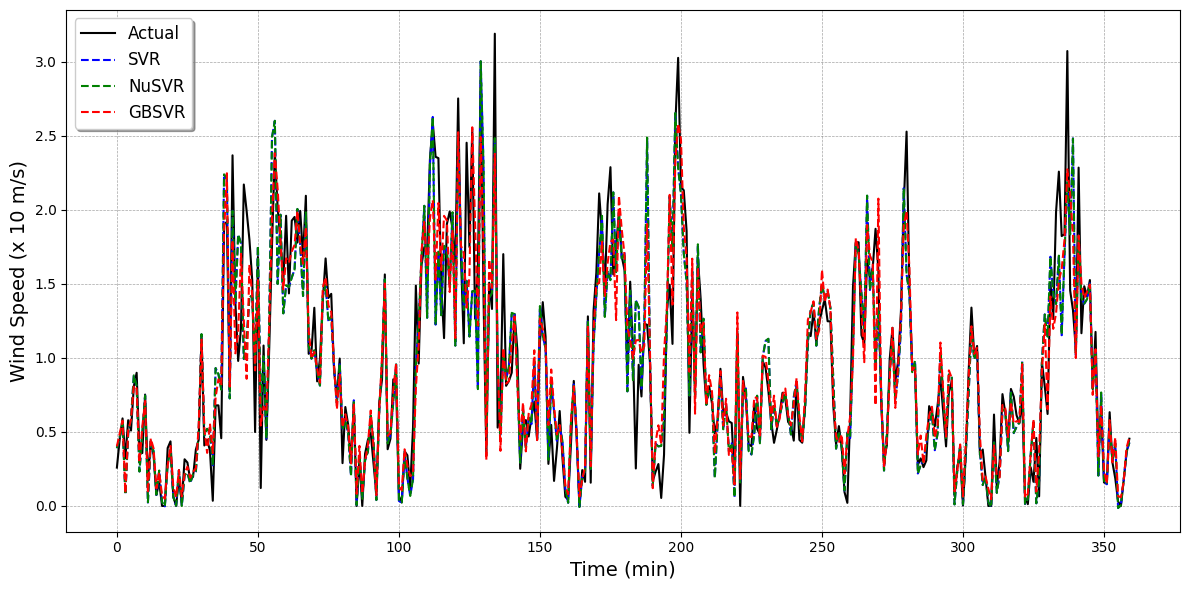}
        \text{20 min interval}
    \end{minipage}
    \hfill
    \begin{minipage}{0.45\textwidth}
        \centering
        \includegraphics[width=\linewidth]{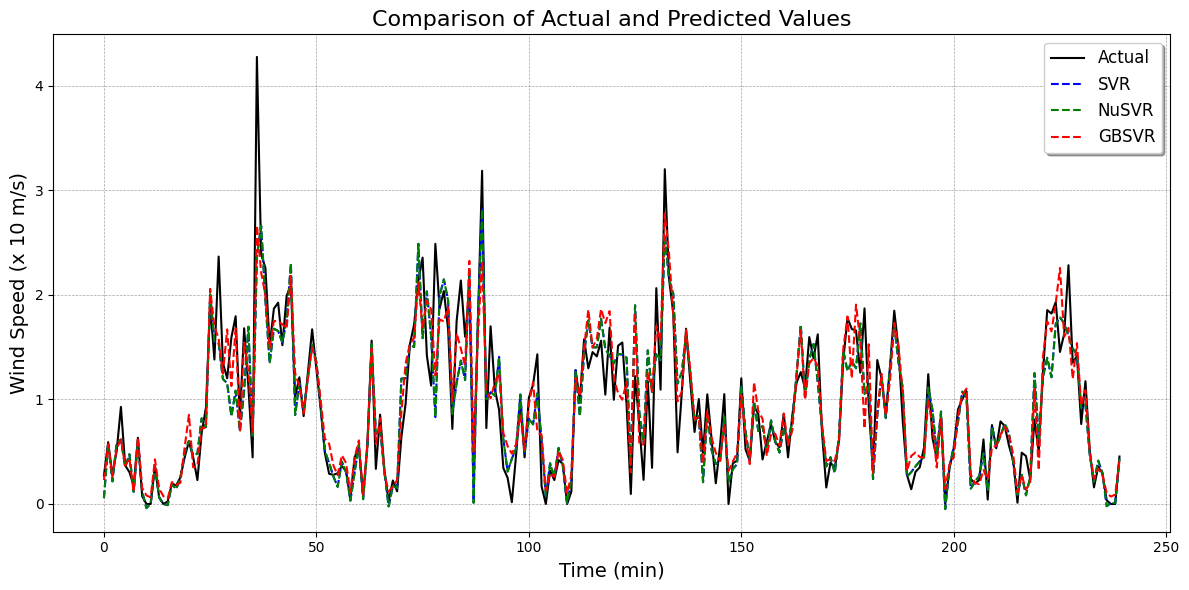}
        \text{30 min interval}
    \end{minipage}
    
    \caption{Wind Speed Prediction}
    \label{wind_speed_graphs}
\end{figure}

The output value at time step \(i\) is \(\vec{y_i} = x_i\), with the sliding window mechanism enabling the model to forecast wind speed at 20-minute and 30-minute intervals. Table \ref{wind_table} compares the performance of GBSVR, SVR, and NuSVR for wind speed prediction across different metrics. Fig. \ref{wind_speed_graphs} shows actual and predicted wind speeds for both intervals. Results demonstrate that GBSVR consistently outperforms SVR and NuSVR, making it highly effective for both short-term and slightly longer-term wind speed forecasting with high accuracy and low error rates.
\begin{table}[ht]
\scriptsize
\centering
\caption{Comparison results on wind dataset at 30-minute and 20-minute intervals.}
\renewcommand{\arraystretch}{1.2}
\setlength{\tabcolsep}{4pt}
\begin{tabular}{|l|cccc|cccc|}
\hline
\multicolumn{5}{|c|}{\textbf{30-Minute Interval}} & \multicolumn{4}{c|}{\textbf{20-Minute Interval}} \\
\hline
\textbf{Model} & \textbf{R$^2$} & \textbf{MAE} & \textbf{MSE} & \textbf{RMSE} & \textbf{R$^2$} & \textbf{MAE} & \textbf{MSE} & \textbf{RMSE} \\
\hline
GBSVR & \textbf{0.8241} & \textbf{0.1896} & \textbf{0.0851} & \textbf{0.2917} & \textbf{0.8502} & \textbf{0.1721} & \textbf{0.0693} & \textbf{0.2633} \\
SVR               & 0.8006 & 0.1957 & 0.0965 & 0.3106 & 0.8091 & 0.1864 & 0.0884 & 0.2973 \\
NuSVR             & 0.8003 & 0.1959 & 0.0966 & 0.3108 & 0.8092 & 0.1863 & 0.0883 & 0.2972 \\
\hline
\end{tabular}
\label{wind_table}
\end{table}


\section{Ablation Study}
This section presents an ablation study on the impact of two key hyperparameters: purity threshold and minimum points per granular regression ball, on model performance using the machines dataset. Performance is evaluated using \( R^2 \) and MAE.

\begin{figure}[ht]
    \centering
    \begin{minipage}{0.35\textwidth}
        \centering
        \includegraphics[width=\linewidth]{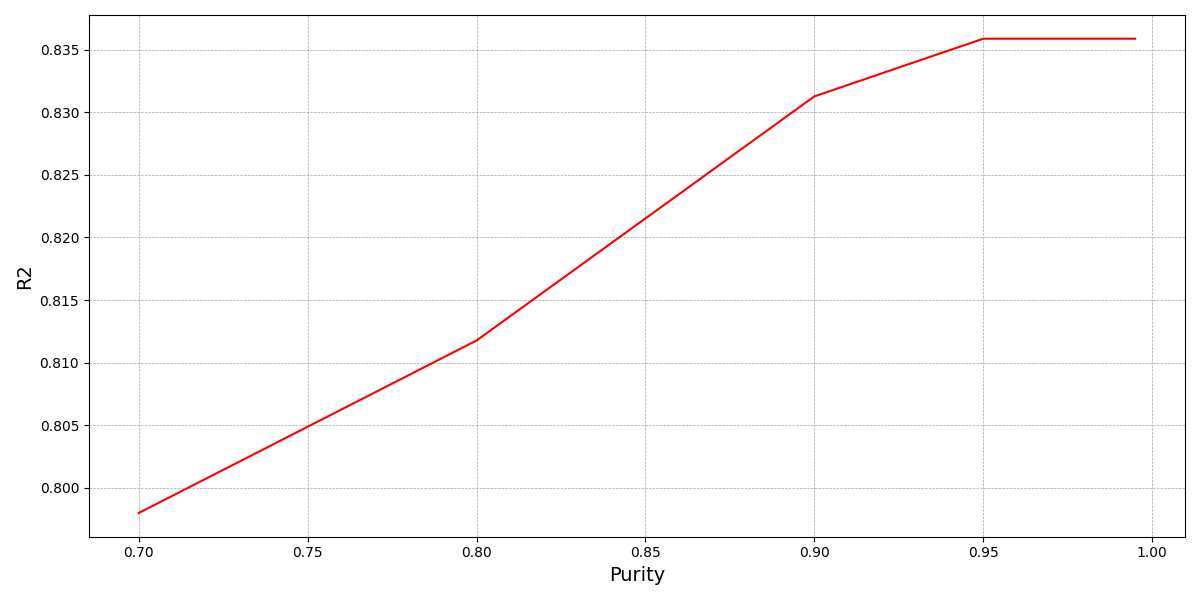}
        \text{$R^2$}
    \end{minipage}
    \hfill
    \begin{minipage}{0.35\textwidth}
        \centering
        \includegraphics[width=\linewidth]{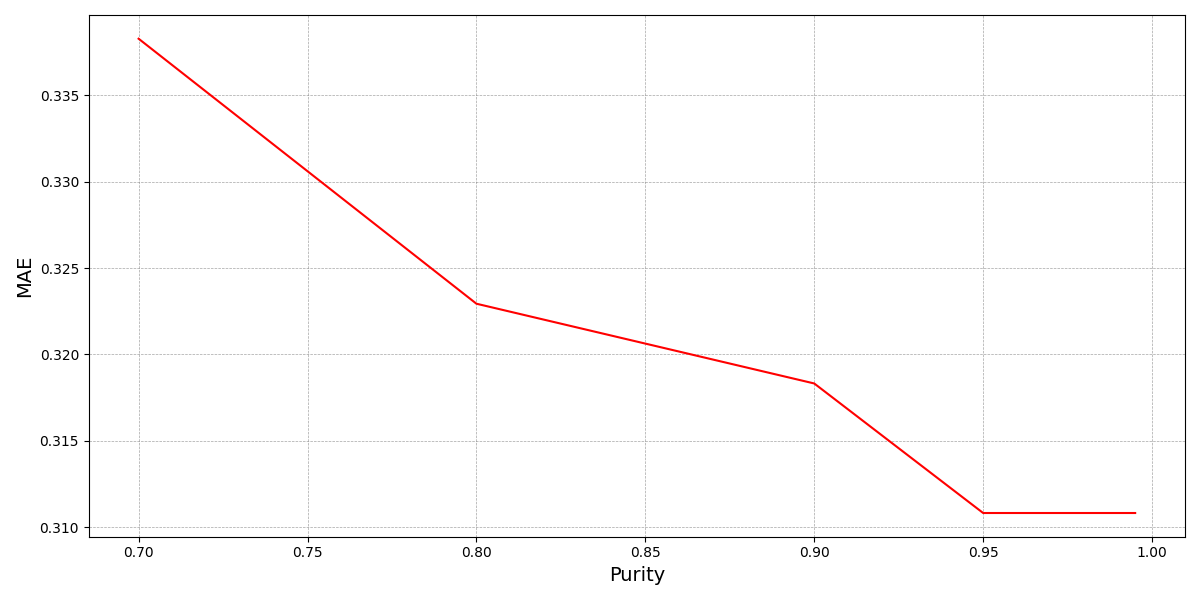}
        \text{MAE}
    \end{minipage}    
    \caption{Ablation Studies on Purity}
    \label{ablation_purity}
\end{figure}


\begin{figure}[ht]
    \centering
    \begin{minipage}{0.35\textwidth}
        \centering
        \includegraphics[width=\linewidth]{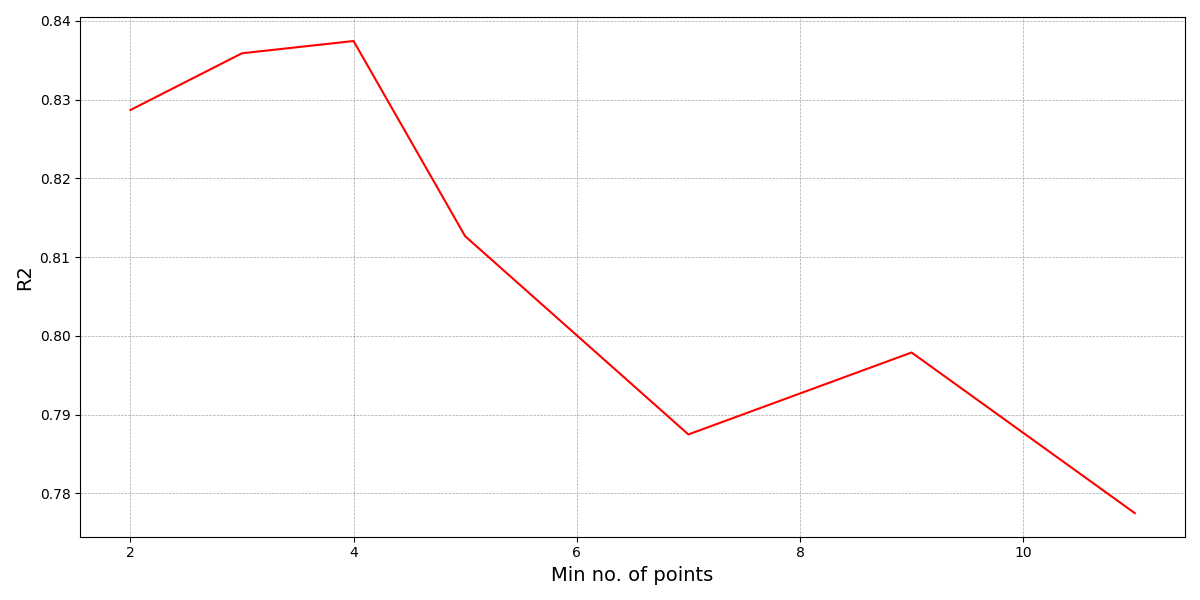}
        \text{$R^2$}
    \end{minipage}
    \hfill
    \begin{minipage}{0.35\textwidth}
        \centering
        \includegraphics[width=\linewidth]{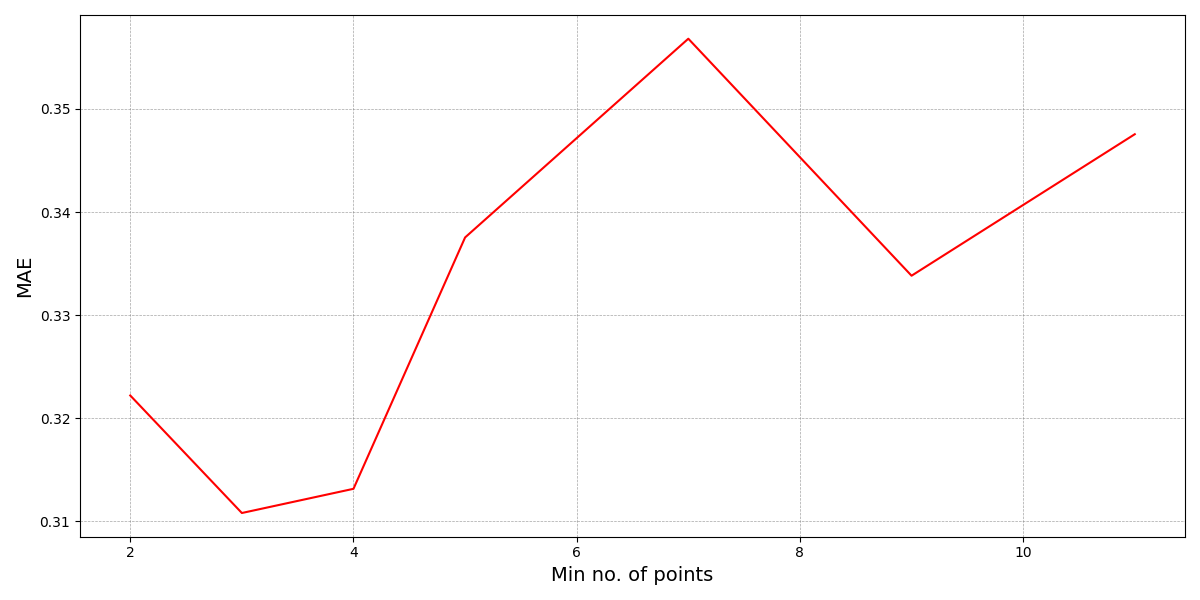}
        \text{MAE}
    \end{minipage}
    \caption{Ablation Studies on Minimum Number of Points in Each Ball}
    \label{ablation_minPts}
\end{figure}

\subsection{Purity}
The impact of purity threshold variation (70\% to 99\%) on model performance was investigated.  The results, presented in Fig. \ref{ablation_purity}, demonstrate a positive correlation between purity threshold and model efficacy.  This observed improvement can be attributed to the reduction of noise within each constructed "ball" as purity increases.  Specifically, a ball generated with a 70\% purity threshold is susceptible to a higher concentration of noisy data points compared to a ball derived from a 99\% threshold.  The application of a more stringent purity criterion results in the iterative subdivision of balls into smaller, more homogeneous clusters, thereby mitigating the associated loss within each individual ball.

\subsection{Granular Regression Ball Cardinality Threshold}
The effect of the minimum cardinality (number of data points) within each ball on model performance was evaluated. Fig. \ref{ablation_minPts} shows that performance improves with increasing minimum points up to an optimal value, after which it declines. This decline occurs due to higher loss as more points are added, leading to reduced homogeneity within each ball. For the Machine dataset, the optimal minimum number of points per ball is 4; deviations from this value result in increased loss and decreased performance.

\subsection{ConMGSVR vs GBSVR}
\begin{figure}[ht]
    \centering
    \includegraphics[width=0.8\linewidth]{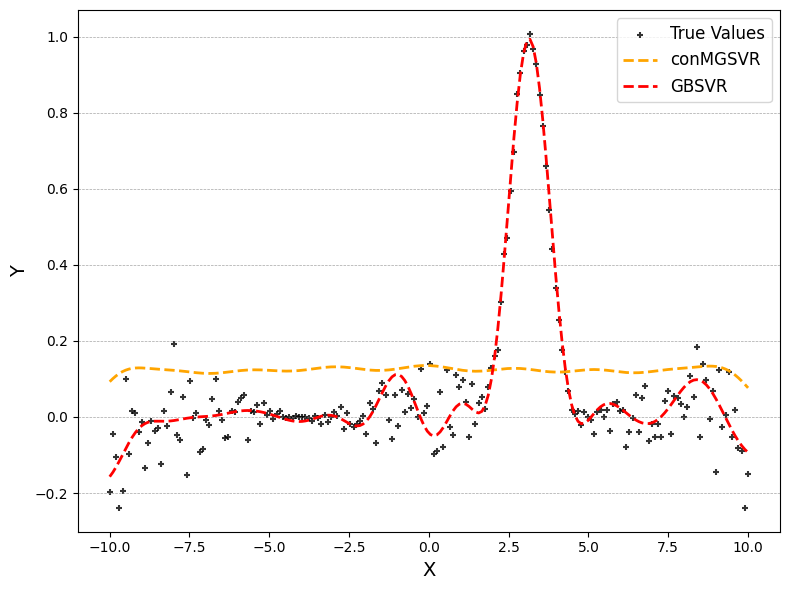} 
    \caption{Visualization of $-cos(x) \exp( -(x - \pi )^2 )$}
    \label{fig:conMGSVR_output}
\end{figure}

\begin{figure}[ht]
    \centering
    \includegraphics[width=0.8\linewidth]{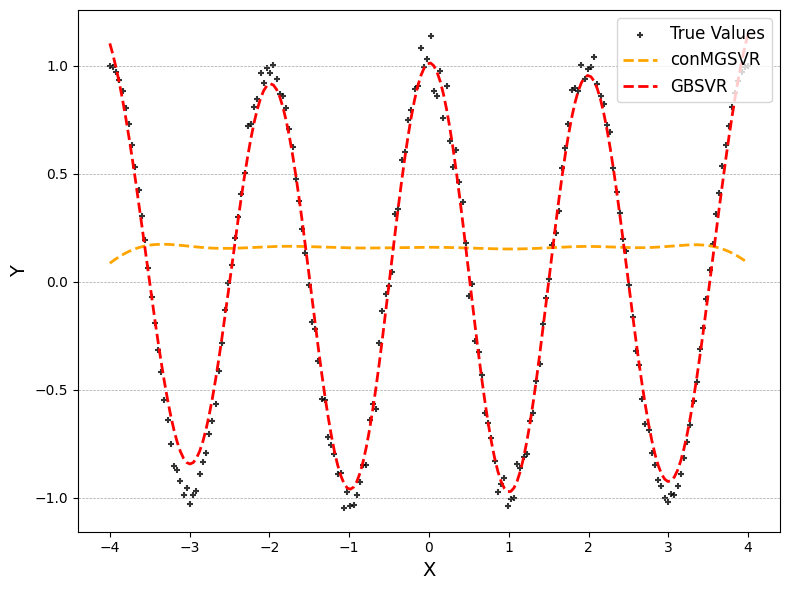} 
    \caption{Visualization of $\cos(\pi x)$}
    \label{fig:conMGSVR_cos}
\end{figure}

To compare GBSVR and the Controllable Multigranularity Support Vector Regression (con-MGSVR) \cite{shao2025mgsvm}, we implemented the dual formulation of con-MGSVR and applied both methods to two artificial datasets: $f(x) = \cos(x)\exp(-{(x-\pi)}^2)$, and $f(x) = \cos(\pi x)$. The results are depicted in Fig. \ref{fig:conMGSVR_output} and Fig. \ref{fig:conMGSVR_cos} respectively.

Results indicate con-MGSVR fails to adequately capture the nonlinear patterns, particularly in regions with sharp variations, due to omission of regression values $y_i$ during the granular regression ball construction process. In contrast, GBSVR captures all nonlinearities, demonstrating superior predictive performance. These results validate our approach of integrating regression values and refining the dual formulation for a more accurate model.


\section{Conclusions}
In this proposed work, we introduced Granular Ball Support Vector Regression (GBSVR), an efficient and novel approach aimed at reducing computational costs and improving robustness to outliers in traditional Support Vector Regression (SVR). Using the concept of granular regression balls, GBSVR reduces computational complexity by summarizing data instances in large datasets into fewer granular representations constructed via discretization method for continuous-valued attributes, facilitating efficient granular regression ball construction.
The mathematical framework of GBSVR 
is proposed and was evaluated on benchmark, artificial, and real domain datasets, demonstrating superior performance compared to existing state-of-the-art regression techniques. These results highlight the effectiveness of GBSVR in maintaining accuracy while significantly reducing computational overhead, as is evident from the GBSVR model training time. Furthermore, the use of granular regression balls provides robustness to outliers, addressing one of the critical limitations of SVR.
This study presents a new approach to handle regression tasks, highlighting the benefit of granular computing in machine learning. Future research could explore the extension of granular regression ball-based methods and new discretization techniques to improve existing work and develop better regression solutions in very large-scale data settings. 

\begin{IEEEbiographynophoto}{Reshma Rastogi} Reshma (nee Khemchandani) Rastogi (M’17) is currently an Associate Professor, Department of Computer Science, South Asian University, India. She has published over 60 papers in international journals and over 35 papers in international conferences in Machine Learning. She is co-author of  TWSVM has been cited over 1800 times and has been a subject of review articles in multiple highly reputed journals. She has co-authored two books including Twin Support Vector Machines: Models, Extensions and Applications (Springer) and Financial Mathematics: An Introduction. Her current research interests include machine learning, image processing, financial modeling, quantum computing and optimization. She is an Action Editor of Neural Networks.
\end{IEEEbiographynophoto}
\vspace{-8mm}
\begin{IEEEbiographynophoto}
{Ankush Bisht} Ankush Bisht is currently pursuing a Master of Science degree at South Asian University. His research focuses on classification and regression techniques, as well as the application of Granular Balls to various machine learning tasks. He has multiple papers accepted in refereed international conferences in the field of machine learning. 
\end{IEEEbiographynophoto}
\vspace{-8mm}
\begin{IEEEbiographynophoto}{Sanjay Kumar} Sanjay Kumar is an Assistant Professor at Department of Computer Science, Deshbandhu College, University of Delhi. His expertise is in  machine learning in particular multi-label learning with missing labels with multiple publications in refereed journals and conferences. 
\end{IEEEbiographynophoto}
\vspace{-8mm}
\begin{IEEEbiographynophoto}{Suresh Chandra} Suresh Chandra is Emeritus Professor of Department of Mathematics,  Indian Institute of Technology Delhi.  He is a Senior Member of the Operational Research Society of India, and a member of the International Working Group on Generalized Convexity and Applications. 
His research interests include numerical optimization, mathematical programming, generalized convexity, fuzzy optimization, fuzzy games, neural networks, machine learning, and financial mathematics. He has authored and coauthored over 150 publications in refereed journals and international conferences with more than 4000 citations. He has also co-authored the widely acclaimed books Numerical Optimization With Applications, Principles of Optimization Theory, Financial Mathematics: An Introduction, Twin Support Vector Machines: Models, Extensions and Applications (Springer), Fuzzy Mathematical Programming and Fuzzy Matrix Games (Springer), Fuzzy Portfolio Optimization (Springer), and Fuzzy Sets and Applications: Logic Modelling and Decision Making.  Prof. Chandra is currently an Editorial Board Member of the journal OPSEARCH (Springer).  Presently he is working on a Research Monograph on "Mathematics Behind Machine Learning".
\end{IEEEbiographynophoto}

\vfill


\begin{thebibliography}{1}
\bibliographystyle{IEEEtran}

\bibitem{darlington2016regression}
R.~B. Darlington and A.~F. Hayes, \emph{Regression analysis and linear models: Concepts, applications, and implementation}.\hskip 1em plus 0.5em minus 0.4em\relax Guilford Publications, 2016.

\bibitem{gunst2018regression}
R.~F. Gunst and R.~L. Mason, \emph{Regression analysis and its application: a data-oriented approach}.\hskip 1em plus 0.5em minus 0.4em\relax CRC Press, 2018.

\bibitem{zaki2020data}
M.~J. Zaki, W.~Meira~Jr, and W.~Meira, \emph{Data mining and machine learning: Fundamental concepts and algorithms}.\hskip 1em plus 0.5em minus 0.4em\relax Cambridge University Press, 2020.

\bibitem{o2016statistical}
C.~M. O'Brien, ``Statistical learning with sparsity: the lasso and generalizations,'' 2016.

\bibitem{drucker1996support}
H.~Drucker, C.~J. Burges, L.~Kaufman, A.~Smola, and V.~Vapnik, ``Support vector regression machines,'' \emph{Advances in neural information processing systems}, vol.~9, 1996.

\bibitem{muchtadi2024support}
I.~Muchtadi-Alamsyah, R.~Viltoriano, F.~Harjono, M.~Nazaretha, M.~Susilo, A.~Bayu, B.~Josaphat, A.~Hakim, and K.~Syuhada, ``Support vector regression-based heteroscedastic models for cryptocurrency risk forecasting,'' \emph{Applied Soft Computing}, p. 111792, 2024.

\bibitem{bi2003geometric}
J.~Bi and K.~P. Bennett, ``A geometric approach to support vector regression,'' \emph{Neurocomputing}, vol.~55, no. 1-2, pp. 79--108, 2003.

\bibitem{vapnik2013nature}
V.~Vapnik, \emph{The nature of statistical learning theory}.\hskip 1em plus 0.5em minus 0.4em\relax Springer science \& business media, 2013.

\bibitem{rastogi2017norm1}
R.~Rastogi, P.~Anand, and S.~Chandra, ``$l_1$-norm twin support vector machine-based regression,'' \emph{Optimization}, vol.~66, no.~11, pp. 1895--1911, 2017.

\bibitem{rastogi2017nu}
{R. Rastogi}, {P. Anand}, and {S. Chandra}, ``A $\nu$-twin support vector machine based regression with automatic accuracy control,'' \emph{Applied Intelligence}, vol.~46, pp. 670--683, 2017.

\bibitem{zhang2024multi}
Z.~Zhang, W.-C. Hong, and Y.~Dong, ``Multi-hyperplane twin support vector regression guided with fuzzy clustering,'' \emph{Information Sciences}, vol. 666, p. 120435, 2024.

\bibitem{gao2019end}
C.~Gao, M.~Shen, X.~Liu, L.~Wang, and M.~Chu, ``End-point static control of basic oxygen furnace (bof) steelmaking based on wavelet transform weighted twin support vector regression,'' \emph{Complexity}, vol. 2019, no.~1, p. 7408725, 2019.

\bibitem{gu2023incremental}
B.~Gu, J.~Cao, F.~Pan, and W.~Xiong, ``Incremental learning for lagrangian $\varepsilon$-twin support vector regression,'' \emph{Soft Computing}, vol.~27, no.~9, pp. 5357--5375, 2023.

\bibitem{yang2025flexible}
M.~Yang, H.~Liang, X.~Wu, and Z.~Zhang, ``A flexible and efficient algorithm for high dimensional support vector regression,'' \emph{Neurocomputing}, vol. 611, p. 128671, 2025.

\bibitem{chuang2002robust}
C.-C. Chuang, S.-F. Su, J.-T. Jeng, and C.-C. Hsiao, ``Robust support vector regression networks for function approximation with outliers,'' \emph{IEEE Transactions on Neural Networks}, vol.~13, no.~6, pp. 1322--1330, 2002.

\bibitem{zadeh2000fuzzy}
L.~A. Zadeh, \emph{Fuzzy Sets and Fuzzy Information-Granulation Theory: Key Selected Papers by}.\hskip 1em plus 0.5em minus 0.4em\relax Beijing Normal University Press, 2000.

\bibitem{xia2019granular}
S.~Xia, Y.~Liu, X.~Ding, G.~Wang, H.~Yu, and Y.~Luo, ``Granular ball computing classifiers for efficient, scalable and robust learning,'' \emph{Information Sciences}, vol. 483, pp. 136--152, 2019.

\bibitem{xie2023efficient}
J.~Xie, W.~Kong, S.~Xia, G.~Wang, and X.~Gao, ``An efficient spectral clustering algorithm based on granular-ball,'' \emph{IEEE Transactions on Knowledge and Data Engineering}, vol.~35, no.~9, pp. 9743--9753, 2023.

\bibitem{xia2024gbsvm}
S.~Xia, X.~Lian, G.~Wang, X.~Gao, J.~Chen, and X.~Peng, ``Gbsvm: an efficient and robust support vector machine framework via granular-ball computing,'' \emph{IEEE Transactions on Neural Networks and Learning Systems}, 2024.

\bibitem{xia2024granular}
S.~Xia, X.~Lian, G.~Wang, X.~Gao, Q.~Hu, and Y.~Shao, ``Granular-ball fuzzy set and its implement in svm,'' \emph{IEEE Transactions on Knowledge and Data Engineering}, 2024.

\bibitem{bai2023granular}
H.~Bai, F.~Shen, W.~Kong, and J.~Feng, ``Granular-ball clustering based neighbourhood outliers detection method,'' in \emph{2023 6th International Conference on Electronics Technology (ICET)}.\hskip 1em plus 0.5em minus 0.4em\relax IEEE, 2023, pp. 1306--1312.

\bibitem{xia2021granular}
S.~Xia, S.~Zheng, G.~Wang, X.~Gao, and B.~Wang, ``Granular ball sampling for noisy label classification or imbalanced classification,'' \emph{IEEE Transactions on Neural Networks and Learning Systems}, vol.~34, no.~4, pp. 2144--2155, 2021.

\bibitem{yong2004improved}
Q.~Yong, Y.~Jie, Y.~Lixiu, and Y.~Chenzhou, ``An improved way to make large-scale svr learning practical,'' \emph{EURASIP Journal on Advances in Signal Processing}, vol. 2004, pp. 1--7, 2004.

\bibitem{zhou2013study}
X.~Zhou and Y.~Ma, ``A study on smo algorithm for solving epsilon-svr with non-psd kernels,'' \emph{Communications in Statistics-Simulation and Computation}, vol.~42, no.~10, pp. 2175--2196, 2013.

\bibitem{sabzekar2021robust}
M.~Sabzekar and S.~M.~H. Hasheminejad, ``Robust regression using support vector regressions,'' \emph{Chaos, Solitons \& Fractals}, vol. 144, p. 110738, 2021.

\bibitem{ye2020robust}
Y.~Ye, J.~Gao, Y.~Shao, C.~Li, Y.~Jin, and X.~Hua, ``Robust support vector regression with generic quadratic nonconvex $\varepsilon$-insensitive loss,'' \emph{Applied Mathematical Modelling}, vol.~82, pp. 235--251, 2020.

\bibitem{fu2023robust}
S.~Fu, Y.~Tian, and L.~Tang, ``Robust regression under the general framework of bounded loss functions,'' \emph{European Journal of Operational Research}, vol. 310, no.~3, pp. 1325--1339, 2023.

\bibitem{chen1982topological}
L.~Chen, ``Topological structure in visual perception,'' \emph{Science}, vol. 218, no. 4573, pp. 699--700, 1982.

\bibitem{guo2021trend}
H.~Guo, L.~Wang, X.~Liu, and W.~Pedrycz, ``Trend-based granular representation of time series and its application in clustering,'' \emph{IEEE Transactions on Cybernetics}, vol.~52, no.~9, pp. 9101--9110, 2021.

\bibitem{kong2022research}
W.~Kong, Y.~Wu, J.~Qi, and Y.~Chen, ``Research on classification label denoising algorithm based on granular ball,'' in \emph{2022 7th International Conference on Cloud Computing and Big Data Analytics (ICCCBDA)}.\hskip 1em plus 0.5em minus 0.4em\relax IEEE, 2022, pp. 102--106.

\bibitem{xia2022efficient}
S.~Xia, X.~Dai, G.~Wang, X.~Gao, and E.~Giem, ``An efficient and adaptive granular-ball generation method in classification problem,'' \emph{IEEE Transactions on Neural Networks and Learning Systems}, vol.~35, no.~4, pp. 5319--5331, 2022.

\bibitem{xia2023granular}
S.~Xia, G.~Wang, X.~Gao, and X.~Lian, ``Granular-ball computing: an efficient, robust, and interpretable adaptive multi-granularity representation and computation method,'' \emph{arXiv preprint arXiv:2304.11171}, 2023.

\bibitem{hui2016heuristic}
Y.~Hui, S.~Wenzhu, Z.~Xiuzhi, Z.~Guotao, and H.~Wenting, ``Heuristic sample reduction based support vector regression method,'' in \emph{2016 IEEE international conference on mechatronics and automation}.\hskip 1em plus 0.5em minus 0.4em\relax IEEE, 2016, pp. 2065--2069.

\bibitem{le2017approximation}
T.~Le, T.~D. Nguyen, V.~Nguyen, and D.~Phung, ``Approximation vector machines for large-scale online learning,'' \emph{Journal of Machine Learning Research}, vol.~18, no. 111, pp. 1--55, 2017.

\bibitem{shieh2010reduced}
H.-L. Shieh and C.-C. Kuo, ``A reduced data set method for support vector regression,'' \emph{Expert Systems with Applications}, vol.~37, no.~12, pp. 7781--7787, 2010.

\bibitem{li2016fast}
Z.~Li, T.~Yang, L.~Zhang, and R.~Jin, ``Fast and accurate refined nystr{\"o}m-based kernel svm,'' in \emph{Proceedings of the AAAI Conference on Artificial Intelligence}, vol.~30, no.~1, 2016.

\bibitem{shao2025mgsvm}
Y.~Shao, Y.~Hua, Z.~Gong, X.~Zhu, Y.~Cheng, L.~Li, and S.~Xia, ``Con-mgsvm: Controllable multi-granularity support vector algorithm for classification and regression,'' \emph{Information Fusion}, vol. 117, p. 102867, 2025.

\bibitem{smola2004tutorial}
A.~J. Smola and B.~Sch{\"o}lkopf, ``A tutorial on support vector regression,'' \emph{Statistics and computing}, vol.~14, pp. 199--222, 2004.

\bibitem{williamson2000new}
R.~Williamson, P.~Bartlett \emph{et~al.}, ``New support vector algorithms,'' \emph{Neural Computation}, vol.~12, pp. 1207--1245, 2000.

\bibitem{UCI_ML_Repository}
M.~Kelly, R.~Longjohn, and K.~Nottingham, ``The uci machine learning repository,'' 2025, accessed: 2025-02-28. [Online]. Available: \url{https://archive.ics.uci.edu}









\end{thebibliography}
\end{document}